\pdfoutput=1

\documentclass[11pt]{article}
\usepackage[dvipsnames,table]{xcolor}

\usepackage{ACL2023}

\usepackage{times}
\usepackage{latexsym}

\usepackage[T1]{fontenc}

\usepackage[utf8]{inputenc}

\usepackage{microtype}

\usepackage{inconsolata}

\usepackage{times}
\usepackage{array}
\usepackage{arydshln}

\usepackage[flushleft]{threeparttable}

\usepackage{hyperref}

\usepackage{relsize}
\usepackage{tabularx}
\usepackage{multirow}
\usepackage{enumitem}
\usepackage{fancyvrb}
\usepackage{tikz}
\usetikzlibrary{patterns}
\usetikzlibrary{positioning}
\usepackage{diagbox}
\usepackage{graphicx}
\usepackage{caption}
\usepackage{subcaption}
\usepackage{bm}
\usepackage{url}
\usepackage{rotating}
\usepackage{pdflscape}
\usepackage{comment}
\usepackage{amssymb}
\usepackage{stfloats}
\usepackage{tabu}
\usepackage{bbm}

\usepackage[normalem]{ulem}

\usepackage{pifont}
\usepackage{siunitx}

\usepackage{float}
 
\usepackage{soul}

\usepackage{amsmath}
\usepackage[export]{adjustbox}

\usepackage{booktabs, makecell, multirow}

\newcommand{\metricname}{\textsc{Lens}}
\newcommand{\frameworkname}{\textsc{Rank \& Rate}}

\newcommand{\datasetname}{\textsc{SimpEval}}
\newcommand{\datasetnameFirst}{\textsc{SimpEval}\textsubscript{\textsc{PAST}}}
\newcommand{\datasetnameSecond}{\textsc{SimpEval}\textsubscript{\textsc{2022}}}
\newcommand{\likertdatasetname}{\textsc{SimpLikert}\textsubscript{\textsc{2022}}}
\newcommand{\dadatasetname}{\textsc{SimpDA}\textsubscript{\textsc{2022}}}
\newcommand{\wikida}{\textsc{Wiki-DA}}
\newcommand{\newlikert}{\textsc{Newsela-Likert}}

\newcommand\modelfont[1]{\smash{{\usefont{T1}{}{m}{n}#1}}}

\definecolor{teal1}{HTML}{d1eeea}
\definecolor{teal2}{HTML}{a8dbd9}
\definecolor{teal3}{HTML}{85c4c9}
\definecolor{teal4}{HTML}{68abb8}
\definecolor{teal5}{HTML}{4f90a6}
\definecolor{teal6}{HTML}{3b738f}
\definecolor{teal7}{HTML}{2a5674}

\definecolor{del}{HTML}{ed8f84}
\definecolor{para}{HTML}{7da3f7}
\definecolor{split}{HTML}{f5cf6c}

\newcommand{\mathlarge}[1]{\mathlarger{\mathlarger{\mathlarger{#1}}}}

\title{\metricname \raisebox{-0.1\height}{\includegraphics[width=1.13em]{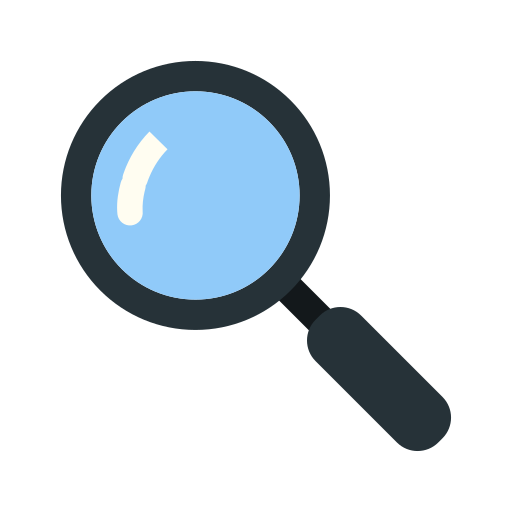}}: A Learnable Evaluation Metric for Text Simplification}

\definecolor{porject_url_color}{HTML}{36a6d6}
\newcommand{\changeurlcolor}[1]{\hypersetup{urlcolor=#1}}  

\author{Mounica Maddela\thanks{\hspace{4pt}Equal contribution.},~ Yao Dou\footnotemark[1],~ David Heineman, Wei Xu \\
School of Interactive Computing \\
  Georgia Institute of Technology \\
  {\small \texttt{\{mmaddela3, douy, david.heineman\}@gatech.edu; wei.xu@cc.gatech.edu}} \\ \\
  \changeurlcolor{porject_url_color}\url{http://lens-score.com/}\\}
  
\begin{document}
\maketitle

\begin{abstract}

Training learnable metrics using modern language models has recently emerged as a promising method for the automatic evaluation of machine translation. However, existing human evaluation datasets for text simplification have limited annotations that are based on unitary or outdated models, making them unsuitable for this approach. To address these issues, we introduce the \datasetname~corpus that contains: \datasetnameFirst, comprising 12K human ratings on 2.4K simplifications of 24 past systems, and \datasetnameSecond, a challenging simplification benchmark consisting of over 1K human ratings of 360 simplifications including GPT-3.5 generated text. Training on \datasetname, we present \metricname, a \textbf{L}earnable \textbf{E}valuatio\textbf{n} Metric for Text \textbf{S}implification. Extensive empirical results show that \metricname~correlates much better with human judgment than existing metrics, paving the way for future progress in the evaluation of text simplification. We also introduce \frameworkname, a human evaluation framework that rates simplifications from several models in a list-wise manner using an interactive interface, which ensures both consistency and accuracy in the evaluation process and is used to create the \datasetname~datasets. 

\end{abstract}

\section{Introduction}
\label{sec:introduction}

\begin{figure}[t!]

\begin{center}
\includegraphics[width=0.99\linewidth]{img/figure1_newest_10x.pdf}
\end{center}
\setlength{\abovecaptionskip}{-1pt}
\setlength{\belowcaptionskip}{-15pt}
\caption{Automatic metric and human evaluation scores on simplifications of a complex sentence from our \datasetnameSecond. \metricname~achieves the best correlation with humans. SARI penalizes simplifications that deviate from the reference (first two examples), and BERTScore fails to penalize hallucinations with high lexical overlap (third example).} 
\label{fig:figure_1}
\end{figure}

Text simplification is a text-to-text generation task that aims to make a text easier to read while preserving its original meaning \cite{Saiggon}. Automatic evaluation of text simplification is challenging because a sentence can be simplified in many ways, such as paraphrasing complex words, deleting insignificant information, and splitting long sentences into shorter ones. An ideal automatic metric should accommodate these diverse choices while capturing semantic similarity and fluency.
However, existing metrics such as SARI \cite{xu-etal-2016-optimizing} and BERTScore \cite{zhang2020bertscore} struggle to capture all the aspects and achieve a high correlation with human evaluation \cite{alva-manchego-etal-2021-un} (see Figure \ref{fig:figure_1}). These metrics fail even more when evaluating high-quality systems that have close performance, calling for a more robust and accurate metric for text simplification.

Prior work on machine translation \cite{rei-etal-2020-comet,sellam-etal-2020-bleurt} has seen the success of using language models as an automatic metric by training them on human judgments, such as Direct Assessment (DA) that rates generations on a 0-100 scale.

However, existing human evaluation datasets \cite{alva-manchego-etal-2021-un, sulem-etal-2018-simple}  for text simplification are not suitable for training metrics because they include a limited number of annotations or systems. Besides, these datasets do not include state-of-the-art generation models such as GPT-3.5 \cite{ouyang2022training} that have been shown to generate human-like quality text \cite{dou-etal-2022-gpt,goyal2022news}.

In this work, we introduce \metricname, a \textbf{L}earnable \textbf{E}valuatio\textbf{n} Metric for Text \textbf{S}implification. \metricname~ is the first supervised metric for the task and uses an adaptive ranking loss to promote fair comparison of simplifications that have undergone different edits (i.e., splitting, paraphrasing, and deletion). To train \metricname, we collect 12K human judgments on 2.4K simplifications by 24 simplification systems from the literature, which we name as the \datasetnameFirst~dataset. We also create \datasetnameSecond~to evaluate \metricname~and other metrics in a realistic and challenging setting of assessing simplifications by state-of-the-art language models on the more recent, complex, and longer sentences published on Wikipedia after Oct 22nd, 2022. \datasetnameSecond~contains over 1K human ratings on 360 simplifications generated by 4 SOTA models, including GPT-3.5, and 2 humans. 

Empirical experiments show that \metricname~achieves a higher correlation of 0.331 with human ratings on \datasetnameSecond, which is more than twice as high as the correlation scores of 0.112 and 0.149 by BERTScore and SARI, respectively. We further demonstrate that incorporating \metricname~into decoding process, using minimum Bayes risk framework \cite{fernandes-etal-2022-quality}, can directly improve the automatic text simplification system's performance. We expect that our data collection method, including \frameworkname, a list-wise human evaluation interface, can be easily adapted to other text generation tasks.

\section{Background}
\label{sec:background}

\paragraph{Issues of Existing Automatic Metrics.} 

\begin{figure}[t!]
  \centering
  \includegraphics[width=1.0\linewidth]{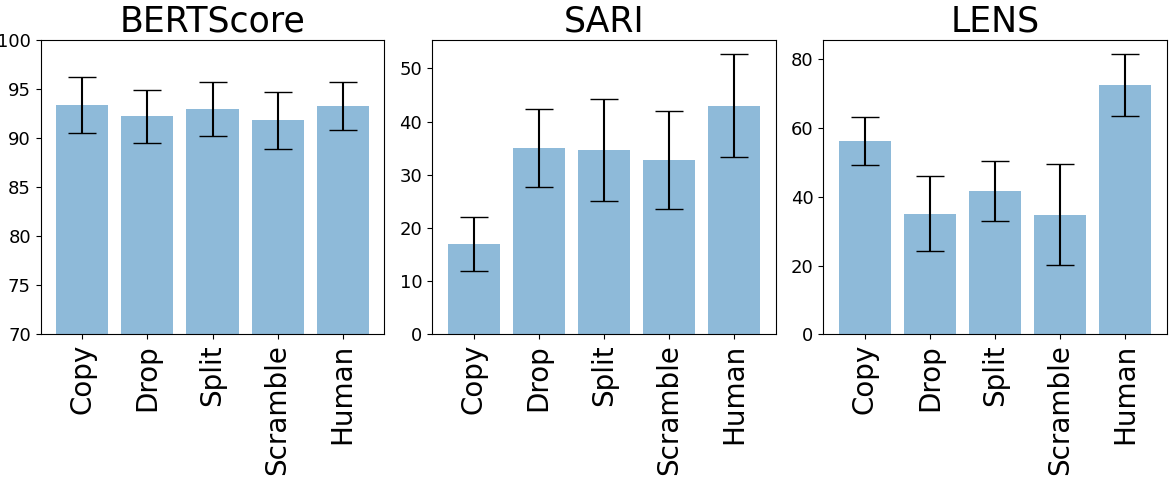}
  \setlength{\abovecaptionskip}{-13pt}
  \setlength{\belowcaptionskip}{-10pt}
  \caption{Average metric scores of conservative (\textit{Copy}) or corrupted (\textit{Drop}, \textit{Scramble}, and \textit{Split}) systems and human (\textit{Human}) simplifications from \datasetname$_{2022}$. The error bar represents the standard deviation of the metric scores. BERTScore ranks some conservative or corrupted systems above \textit{Human}. SARI penalizes \textit{Copy} but ranks the corrupted systems above \textit{Copy}. \metricname~correctly ranks \textit{Human} above the rest and \textit{Copy} above the disfluent corrupted systems.}
  \label{fig:conservative_analysis}
\end{figure}

SARI \cite{xu-etal-2016-optimizing} is the most commonly used metric for text simplification that computes F1/precision scores of the n-grams inserted, deleted, and kept when compared to human references. As SARI measures n-gram string overlap, it penalizes simplifications that are synonymous to the reference but uses different words (see first two examples in Figure \ref{fig:figure_1}). \citet{alva-manchego-etal-2021-un} showed that BERTScore \cite{zhang2020bertscore}, which measures similarity based on BERT \cite{devlin-etal-2019-bert} embeddings, is better at capturing semantic similarity between the simplification system's outputs and the references. However, it fails to penalize conservative systems that make trivial or no changes to the input as illustrated in Figure \ref{fig:conservative_analysis}. This figure shows average metric scores by SARI and BERTScore for a conservative system that copies the input (\textit{Copy}),  an oracle system of human simplification (\textit{Human})\footnote{We use \datasetnameSecond~ (\S \ref{sec:simpeval}) for the analysis. Out of the two human simplifications, we randomly selected one as the oracle and the other as the reference to compute metric scores.}, and three disfluent corrupted systems that drop 10\% of the input words (\textit{Drop}), scramble 5\% of the input words (\textit{Scramble}), or insert a period in the middle (\textit{Split}).  BERTScore ranks some conservative or corrupted systems above human simplifications. SARI penalizes the conservative system because it focuses on the edits performed by the system but ranks the disfluent systems above the conservative system. An ideal automatic metric for text simplification should capture both semantic similarity and the edits performed by the system.

\paragraph{Lack of Human Evaluation Data.}

There is a lack of suitable human evaluation datasets for fine-tuning pre-trained language models for simplification evaluation. \citet{alva-manchego-etal-2021-un} released 0-100 continuous scale ratings on fluency, meaning preservation, and simplicity for only 600 simplifications sampled from six systems. \citet{sulem-etal-2018-simple} released 5-point Likert scale ratings on the same three dimensions for 1,960 simplifications generated by different configurations of six systems. Both datasets are relatively small and cover too few different system designs. For example, they do not include the newer state-of-the-art systems \cite{sheang-saggion-2021-controllable, martin-etal-2022-muss} that are based on T5 \cite{raffel2020exploring} and other large pre-trained language models.

\section{Automatic Evaluation Metric}
\label{sec:lens}

To tackle the challenge of limited human evaluation data, we curate \datasetname, a corpus containing over 13K human judgements on 2.8K simplification from 26 systems. This facilitates the training and evaluation of \metricname~(A \textbf{L}earnable \textbf{E}valuatio\textbf{n} Metric for Text \textbf{S}implification), the first supervised automatic metric for text simplification evaluation.
In this section, we first describe the creation of \datasetname~datasets in \S\ref{sec:simpeval} and then \metricname~in \S\ref{subsec:lens}.

\subsection{Collecting Human Judgements}
\label{sec:simpeval}
We collect \datasetnameFirst, containing 12K human ratings on 2.4K simplifications from 24 systems on sentences from TurkCorpus \cite{xu-etal-2016-optimizing}, to train \metricname. For evaluating \metricname~and other simplification metrics, we create \datasetnameSecond~that consists of 1,080 human ratings on 360 simplifications from both humans and SOTA models, including GPT-3.5. It features more complex sentences from Wikipedia written after Oct 22nd, 2022, very recent to the time we conduct the experiments, to reduce the risk of ``data contamination'' (i.e., appearing in the training data of LLMs) and serve as a more challenging test bed for large language models. Table \ref{tab:simpeval_stats} shows the summary of both datasets.

\paragraph{A Diverse Set of Simplification Systems.} We consider the following systems (further details in Appendix \ref{app:systems}): (i) two GPT-3.5\footnote{Specifically, we use the \modelfont{text-davinci-003} model which is the most recent variant with 175B parameters. We provide the prompts we used in Appendix \ref{app:gpt3_setup}.} outputs under zero-shot and 5-shot settings; (ii) eight fine-tuned Transformer-based systems of varied sizes and parameter settings \cite{sheang-saggion-2021-controllable, raffel2020exploring, martin-etal-2020-controllable, maddela-etal-2021-controllable}; (iii) three supervised BiLSTM-based systems that use vanilla RNN, reinforcement learning \cite{zhang-lapata-2017-sentence}, or explicit editing \cite{dong-etal-2019-editnts}; (iv) one unsupervised and one semi-supervised system utilizing auto-encoders \cite{surya-etal-2019-unsupervised}; (v) two systems that apply statistical machine translation approaches to simplification \cite{wubben-etal-2012-sentence, xu-etal-2016-optimizing}; (vi) a rule-based system \cite{dhruv-acl-2020}; (vii) three hybrid systems that combine linguistic rules with data-driven methods \cite{narayan-gardent-2014-hybrid, sulem-etal-2018-semantic-short, maddela-etal-2021-controllable}; (viii) two naive baselines that copy the input or scramble 5\% of the input words; and (ix) six human-written simplifications, including two from ASSET \cite{fern2020asset}, one from TurkCorpus \cite{xu-etal-2016-optimizing}, one from Simple Wikipedia that were automatically aligned \cite{kauchak-2013-improving}, and two newly written by our trained in-house annotators.

\begin{table}[t!]
\centering
\setlength{\tabcolsep}{3pt}
\resizebox{\linewidth}{!}{%
{\renewcommand{\arraystretch}{1.12}
\begin{tabular}{lcc>{\centering\arraybackslash}p{1.25cm}>{\centering\arraybackslash}p{1.3cm}}\toprule

\multirow{2}*{\textbf{System}} & \multirow{2}*{\textbf{Arch.}} & \multirow{2}*{\textbf{Data}} & \multicolumn{2}{c}{\textbf{Human Avg.}}   \\ 
& & & \textbf{Raw} & \textbf{Z-Score} \\
\midrule
\multicolumn{5}{l}{\textbf{\datasetnameFirst} \textit{(24 systems on 100 original sentences)}}\\
Human-1 \citeyearpar{fern2020asset} & Human & ASSET & \cellcolor{teal7}\textcolor{white}{86.69} & 
 \cellcolor{teal7}\textcolor{white}{0.783}   \\ 
Human-2 \citeyearpar{fern2020asset} &  Human & ASSET & \cellcolor{teal7}\textcolor{white}{86.12} & \cellcolor{teal7}\textcolor{white}{0.711}  \\ 
MUSS \citeyearpar{martin-etal-2022-muss} & BART-large & \begin{tabular}{@{}c@{}}WikiLarge \\ + Mined Data\end{tabular} & \cellcolor{teal7}\textcolor{white}{84.48}  & \cellcolor{teal7}\textcolor{white}{0.653} \\ 
ControlT5 \citeyearpar{sheang-saggion-2021-controllable} & T5-base & WikiLarge & \cellcolor{teal7}\textcolor{white}{84.70} & \cellcolor{teal7}\textcolor{white}{0.650}  \\
T5-3B \citeyearpar{raffel2020exploring} & T5 & WikiAuto & \cellcolor{teal6}\textcolor{white}{82.79}   & \cellcolor{teal6}\textcolor{white}{0.492} \\ 
T5-large  & T5 & WikiAuto & \cellcolor{teal6}\textcolor{white}{81.86} & \cellcolor{teal6}\textcolor{white}{0.453} \\ 
T5-base & T5 & WikiAuto  & \cellcolor{teal6}\textcolor{white}{82.15} & \cellcolor{teal6}\textcolor{white}{0.443} \\ 
Transformer \citeyearpar{transformer-vaswani-2017}  & BERT-TF & WikiAuto & \cellcolor{teal5}\textcolor{white}{79.42}   & \cellcolor{teal5}\textcolor{white}{0.366}  \\ 
Controllable \citeyearpar{maddela-etal-2021-controllable}  & BERT-TF & WikiAuto  & \cellcolor{teal5}\textcolor{white}{79.44} & \cellcolor{teal5}\textcolor{white}{0.323}   \\ 
Human-3 \citeyearpar{xu-etal-2016-optimizing} & Human & TurkCorpus  & \cellcolor{teal5}\textcolor{white}{79.54} & \cellcolor{teal5}\textcolor{white}{0.281}  \\ 
Human-4 &  Human & Simple Wiki & \cellcolor{teal5}\textcolor{white}{78.36} & \cellcolor{teal5}\textcolor{white}{0.249} \\ 
DRESS \citeyearpar{zhang-lapata-2017-sentence} & BiLSTM+RL & WikiLarge  & \cellcolor{teal5}\textcolor{white}{77.18} & \cellcolor{teal5}\textcolor{white}{0.206} \\ 
Copy & -- & --  & \cellcolor{teal4}{76.81} & \cellcolor{teal4}{0.103}  \\ 
ACCESS \citeyearpar{martin-etal-2020-controllable} & Transformer & WikiLarge  & \cellcolor{teal4}{73.25} & \cellcolor{teal4}{0.001} \\ 
SBMT-SARI \citeyearpar{xu-etal-2016-optimizing} & Statistic MT & PWKP & \cellcolor{teal4}{73.66}  & \cellcolor{teal4}{-0.014} \\ 
PBMT-R  \citeyearpar{wubben-etal-2012-sentence} & Statistic MT & PWKP & \cellcolor{teal4}{72.44} & \cellcolor{teal4}{-0.066} \\ 
EditNTS  \citeyearpar{dong-etal-2019-editnts} & BiLSTM & WikiLarge  & \cellcolor{teal4}{70.15} & \cellcolor{teal4}{-0.162} \\ 
BiLSTM & BiLSTM & WikiLarge  & \cellcolor{teal4}{68.35} & \cellcolor{teal4}{-0.245} \\ 
SEMosses \citeyearpar{sulem-etal-2018-simple} & LSTM & WikiLarge & \cellcolor{teal3}{62.84}  & \cellcolor{teal3}{-0.565} \\ 
UNTS \citeyearpar{surya-etal-2019-unsupervised} & RNN & Wiki dump  & \cellcolor{teal3}{62.66} & \cellcolor{teal3}{-0.596} \\ 
UNMT \citeyearpar{Artetxe2018UnsupervisedNM}  & RNN  & Wiki dump  & \cellcolor{teal3}{60.43}  & \cellcolor{teal3}{-0.673} \\ 
Rule-based \citeyearpar{dhruv-acl-2020} & -- & --  & \cellcolor{teal3}{60.15} & \cellcolor{teal3}{-0.687} \\ 
Hybrid \citeyearpar{narayan-gardent-2014-hybrid} & Statistic MT & PWKP   & \cellcolor{teal2}{55.36} & \cellcolor{teal2}{-0.925} \\ 
Scramble & -- & --  & \cellcolor{teal1}{35.17} & \cellcolor{teal1}{-1.954} \\ \midrule
\multicolumn{5}{l}{\textbf{\datasetnameSecond} \textit{(6 systems on 60 original sentences)}} \\
Human-5  & Human & Ours & \cellcolor{teal7}\textcolor{white}{81.87} & \cellcolor{teal7}\textcolor{white}{0.424}   \\ 
Human-6 & Human & Ours & \cellcolor{teal7}\textcolor{white}{81.57} & \cellcolor{teal7}\textcolor{white}{0.395}   \\ 
GPT-3.5 \citeyearpar{ouyang2022training}  & InstructGPT & 5-shot & \cellcolor{teal6}\textcolor{white}{79.59} & \cellcolor{teal6}\textcolor{white}{0.280}    \\
GPT-3.5 \citeyearpar{ouyang2022training} & InstructGPT & 0-shot & \cellcolor{teal5}\textcolor{white}{75.27} & \cellcolor{teal5}\textcolor{white}{-0.025}  \\ 
MUSS \citeyearpar{martin-etal-2022-muss}  & BART-large & See above & \cellcolor{teal4}{70.74} & \cellcolor{teal4}{-0.333} \\ 
T5-3B \citeyearpar{raffel2020exploring}  & T5 & WikiAuto & \cellcolor{teal3}{64.98} & \cellcolor{teal3}{-0.700} \\ 
\bottomrule
\end{tabular}
}}
\setlength{\belowcaptionskip}{-10pt}
\caption{\datasetnameFirst~and \datasetnameSecond~datasets of human evaluation data that covers a wide range of simplification systems. MUSS is the best-performing model in \datasetnameFirst~with a small gap to humans but is much worse on the more challenging sentences in \datasetnameSecond~where GPT-3.5 performs better. BERT-TF: BERT-base initialized Transformer. Z-scores \cite{graham-etal-2013-continuous,akhbardeh-etal-2021-findings} are standardized based on each rater's mean and standard deviation.}
\label{tab:simpeval_stats}
\end{table}

\vspace{-.2cm}
\noindent \paragraph{Complex Sentences Selection.} 

For \datasetnameFirst, we sample 100 complex sentences from the test set of the widely-used TurkCorpus \cite{xu-etal-2016-optimizing} and ASSET \cite{fern2020asset} evaluation benchmarks, which share the same set of complex sentences but have different human simplifications in terms of edit operations. ASSET contains 10 human references for each complex sentence that are used to train \metricname. 
 GPT is pre-trained on a vast amount of web text, and the sentences in TurkCorpus/ASSET are derived from the 12-year-old Parallel Wikipedia Simplification corpus \cite{zhu-etal-2010-monolingual}, which may have already been encountered by GPT-3.5. To address this ``data contamination'' issue and provide a more challenging benchmark for state-of-the-art models, we manually select 60 long, complex sentences covering recent world events such as the Twitter acquisition and World Cup for \datasetnameSecond.\footnote{For example, \textit{"Musk stated that Twitter Blue's pricing would be raised to around US\$8.00 per-month, and include reduced advertising on the Twitter service, the ability to post longer audio and video files, and verified account status."}} These sentences are from new revisions or articles added to Wikipedia between October 22nd and November 24th, 2022. They have an average length of 38.5 tokens, much longer than the 19.7-token average of TurkCorpus and ASSET.

\vspace{-.2cm}
\noindent \paragraph{Data Annotation.} We ask in-house annotators to rate each simplification on a single 0-100 overall quality scale using our \frameworkname~framework (more details in \S \ref{sec:evaluation_framework}).  We provided an extensive tutorial and two training sessions to the annotators, which involved rating system outputs for five input sentences (screenshots in Appendix \ref{sec:interface}). We periodically inspected the annotations to prevent the deterioration of quality over time. All annotators are English speakers who are university students. Each simplification in \datasetnameFirst~receives 5 ratings, while each simplification in \datasetnameSecond~is rated by 3 annotators. We follow WMT \cite{akhbardeh-etal-2021-findings} to normalize the raw scores by the mean and standard deviation of each rater, also known as the z-score, which are later used to train \metricname. The inter-annotator agreement for \datasetnameFirst~is 0.70 using the interval Krippendorff's $\alpha$ on z-scores, and 0.32 for \datasetnameSecond, partly because it contains GPT-3.5 outputs that are quite competitive with human. Both are considered fair to good agreement \cite{krippendorff2004reliability}.

\subsection{A New Learnable Metric -- \metricname}
\label{subsec:lens}

 Given an input text $c$, the corresponding system output $s$, and a set of $n$ references $R = \{r_1, r_2, \dots, r_n \}$, \metricname~produces a real-valued score $z_{max} = \max_{1 \leq i \leq n}(z_i)$ that maximizes over the quality scores $z_i$ of $s$ in regards to each reference $r_i$. Our model encodes all texts into vectors ($\mathbf{c}, \mathbf{s}, \mathbf{r_i}$) using Transformer-based encoders such as RoBERTa \cite{Liu2019RoBERTaAR}, then combines them into an intermediate representation $\mathbf{H} = \big[\mathbf{s}; \mathbf{r_i}; \mathbf{s}\; \odot\; \mathbf{c}; \mathbf{s}\; \odot\; \mathbf{r_i}; |\mathbf{s}\;-\;\mathbf{c}|; |\mathbf{s}\;-\;\mathbf{r_i}| \big]$ by concatenation ($[;]$) and element-wise product ($\odot$), which is then fed to a feedforward network to predict $z_i$.

\setlength{\tabcolsep}{5pt}
\begin{table*}[ht!]
\small
\centering
\begin{tabular}{lccccccccc}
\toprule 
& \multicolumn{3}{c}{\textbf{\datasetname $_{2022}$}} 
& \multicolumn{3}{c}{\textbf{\wikida}} 
& \multicolumn{3}{c}{\textbf{\newlikert}} \\
\cmidrule(lr){2-4} \cmidrule(lr){5-7} \cmidrule(lr){8-10}
& $\mathbf{\mathlarge{\tau}_{para}}$ & 
$\mathbf{\mathlarge{\tau}_{spl}}$ & $\mathbf{\mathlarge{\tau}_{all}}$ &
\textbf{Fluency} & \textbf{Meaning} & \textbf{Simplicity} & 
\textbf{Fluency} & \textbf{Meaning} & \textbf{Simplicity }\\ 
\midrule
FKGL & -0.556 & -0.31 & -0.356  & 0.054 & 0.145 & 0.001 
 & 0.193 & 0.306 & -0.051 \\
BLEU & 0.048 & -0.054 & -0.033 & 0.460 & 0.622 & 0.438
& 0.332 & 0.261 & 0.118 \\
SARI &  0.206 & 0.140 & 0.149 & 0.335 & 0.534 & 0.366
& 0.234 & 0.124 & 0.094 \\
BERTScore & 0.238 & 0.093 & 0.112 & 0.636 & \textbf{0.682} & 0.614  & 0.384 & 0.274 & 0.215 \\
\midrule
\metricname$_{all}$ & 0.333 & 0.233 & 0.241 & \textbf{0.816} & 0.662 & \underline{0.733} & \textbf{0.655} & \textbf{0.477} & \underline{0.343}\\
\metricname$_{k=1}$ & \textbf{0.460} & \underline{0.295} & \underline{0.307}  & 0.796 & 0.647 & 0.721  & \underline{0.633} & \underline{0.444} & 0.328 \\
\metricname$_{k=3}$ & \underline{ 0.429} & \textbf{0.333} & \textbf{0.331}  & \underline{0.807} & \underline{0.668} & \textbf{0.749} & 0.624 & 0.428 & \textbf{0.359} \\
\bottomrule
\end{tabular}
\caption{Correlation results between automatic metrics and three human ratings datasets: \datasetnameSecond~(this work), \wikida ~\cite{alva-manchego-etal-2021-un}, and \newlikert~\cite{maddela-etal-2021-controllable}. $\mathbf{\mathlarge{\tau}_{para}}$, $\mathbf{\mathlarge{\tau}_{spl}}$, and $\mathbf{\mathlarge{\tau}_{all}}$ represent  the Kendall Tau-like correlation for paraphrase-focused, split-focused, and all simplifications, respectively. We report the Pearson correlation coefficients along three dimensions for \wikida~and \newlikert. The best values are marked in \textbf{bold} and the second best values are \underline{underlined}.}
\label{table:metric_eval_results_simpeval}
\end{table*}

\begin{figure*}[ht!]
  \centering
  \vspace{-8pt}
  \includegraphics[width=1.0\linewidth]{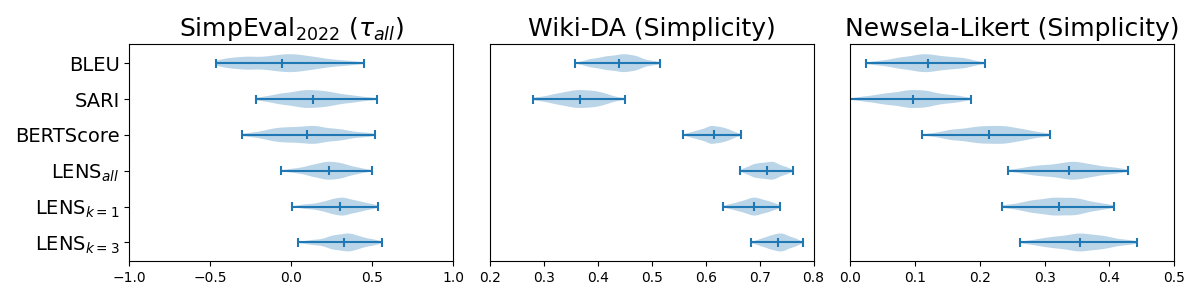}
  \setlength{\abovecaptionskip}{-10pt}
  \setlength{\belowcaptionskip}{-10pt}
  \caption{The 95\% confidence intervals for Kendall Tau-like correlation ($\mathbf{\mathlarge{\tau}_{all}}$) on \datasetnameSecond~ and for Pearson correlation with simplicity ratings on \wikida~and \newlikert, calculated by bootstrapping \cite{deutsch-etal-2021-statistical}. \metricname~is more reliable with smaller intervals and has higher correlation with human judgments.}
  \label{fig:ci_analysis}
\end{figure*}

For training, besides considering all references equally (i.e., \metricname$_{all}$ when $k=n$ in Eq. (\ref{loss})), we also adopt a reference-adaptive loss that selects a subset of references closer to $s$ in terms of edit operations rather than the entire set $R$. It encourages the metric to consider that different simplifications (e.g., paraphrasing-focused, deletion-focused, with or without splitting) can be acceptable, as long as they are close to some (not necessarily all) of the human references. We compute this loss ($L_{adapt}$) as: 
\begin{equation} \label{loss}
L_{adapt} =  \frac{1}{km} \sum_{j=1}^{m} \sum_{z_l \in Z'} (h_j - z_l)^2  \\
\end{equation}
\noindent
where $h$ is human rating and $m$ is the training data size. We compute the set of predicted scores $Z = \{z_1, z_i, \ldots, z_n \}$ corresponding to references in $R$ and then choose top $k$ ($k \leq n$) values from $Z$ to form a subset $Z' \subseteq Z$. Finally, we calculate the mean squared error (MSE) between the human rating $h$ and each score in $Z'$. By selecting top $k$ scores in $Z$, we focus the training of metric on references similar to $s$ in terms of writing style or editing operations. This loss also aligns the training step with the inference step, where we select the best-predicted score $z_{max}$ corresponding to $R$. Although multiple references are ideal, the proposed loss can also use a single reference, similar to the standard MSE loss. We train \metricname~on our \datasetnameFirst~dataset (details in \S \ref{sec:simpeval}). We provide further implementation details in Appendix \ref{app:metric_impl}.

Similar to the COMET \cite{rei-etal-2020-comet} and BLEURT \cite{sellam-etal-2020-bleurt} metrics for MT evaluation, \metricname~is trained on human ratings after z-score normalization, which are real values predominantly lie between [-3, 3]. To make \metricname~more interpretable, we rescale the predicted scores to the range between [0, 100] as the percentage of area under a normal curve of mean $0$ and standard deviation $1$ corresponding to $z_i$. We present the rescaled scores for experiments and analyses in this paper.
Besides, we use RoBERTa-large as the underlying model of \metricname~throughout the paper, unless otherwise specified (\S \ref{sec:lens_analysis}).

\section{Experiments}

\label{sec:eval_results}

We benchmark \metricname~metric against the existing automatic metrics (i) for evaluating text simplification system outputs and (ii) for training better automatic simplification models when used as alternative reward functions.

\subsection{Correlation with Human Evaluation}
We demonstrate that \metricname~correlates better with human judgments than the existing metrics.

\paragraph{Evaluation Datasets.} We compare \metricname~to the existing metrics, namely SARI \cite{xu-etal-2016-optimizing},\footnote{\url{https://github.com/cocoxu/simplification/blob/master/modified-sari.py}} BERTScore \cite{zhang2020bertscore},\footnote{We use precision score of BERTScore as it has been shown to perform better for simplification evaluation \cite{alva-manchego-etal-2021-un}.} BLEU, and Flesch-Kincaid Grade Level readability (FKGL) \cite{kincaid} on three datasets: \datasetnameSecond~(\S \ref{sec:simpeval}), \wikida~released by \citet{alva-manchego-etal-2021-un} with 0-100 continuous scale ratings on fluency, meaning preservation, and simplicity for 600 simplifications across six systems, and \newlikert~collected by \citet{maddela-etal-2021-controllable} with 5-point Likert scale ratings on the same dimensions for 500 simplifications across five systems. While \datasetnameSecond~and \wikida~are derived from Wikipedia, \newlikert~is derived from news articles in Newsela \cite{Xu-EtAl:2015:TACL}, a widely used corpus for text simplification.
When calculating metric scores for \datasetnameSecond~that contains two human simplifications, we use one as the reference and the other as the oracle simplification system. 
We remove the complex sentences in \wikida~that overlap with \datasetnameFirst, the training dataset for \metricname.

\paragraph{Evaluation Setup.}  We report Kendall Tau-like correlation for \datasetnameSecond~to capture the ability of metrics to distinguish two systems, which is close to the real-world scenario. Kendall Tau-like correlation is predominantly used in machine translation metric evaluation at WMT \cite{bojar-etal-2017-results, ma-etal-2018-results} for the same reason as it focuses on the relative ranking of the outputs. Given an input $c$ and its simplifications from $N$ systems $S = \{s_1, \ldots, s_m, \ldots, s_n, \ldots, s_N\}$, we extract $(s_m, s_n)$ pairs, where $1\leq m < n \leq N$, and calculate Kendall Tau-like coefficient $\mathbf{\tau}$:

\begin{equation}
    \tau = \frac{|Concordant| - |Discordant|}{|Concordant| + |Discordant|}
\end{equation}
\noindent
where \textit{Concordant} is the set of pairs where the metric ranked $(s_m, s_n)$ in the same order as humans ranked and \textit{Discordant} is the set of the pairs where the metric and humans disagreed. We report separate coefficients for paraphrase- ($\mathbf{\tau_{para}}$), and split-focused ($\mathbf{\tau_{spl}}$) simplifications, along with ($\mathbf{\tau_{all}}$) for all of them. Following the literature, we only used the $(s_m, s_n)$ pairs for which the difference in ratings is more than 5 out of the 100 points, and all the annotators agreed on the ranking order. To make results comparable to existing work \cite{alva-manchego-etal-2021-un, fern2020asset} that evaluated on \wikida~and \newlikert, we report Pearson correlation ($\rho$) between the metric scores and the human ratings.

\paragraph{Results.} Table \ref{table:metric_eval_results_simpeval} shows that \metricname~outperforms the existing metrics on \datasetnameSecond~ and \wikida~belonging to the Wikipedia domain, and \newlikert~based on newswire domain. The difference is more substantial on \datasetnameSecond~consisting of similar performing SOTA systems, where the $\tau_{all}$ of \metricname$_{k=3}$ exceeds the $\tau_{all}$ of BERTScore by 0.22 points. Training using top $k$ references (\metricname$_{k=3}$) has improved $\tau_{all}$  on \datasetnameSecond~and $\rho$ along the simplicity dimension on the rest when compared to using all the references (\metricname$_{all}$). Figure \ref{fig:ci_analysis} shows the 95\% confidence intervals for $\tau_{all}$ on \datasetnameSecond~and $\rho$ for simplicity, which is deemed to be the most important dimension in prior work \cite{alva-manchego-etal-2021-un, xu-etal-2016-optimizing}, on \wikida~and \newlikert~calculated using bootstrapping methods by \citet{deutsch-etal-2021-statistical}. A smaller interval indicates that the metric exhibits lower variance and higher reliability. \metricname~exhibits smaller intervals than the other metrics.

\subsection{\metricname~as Training or Decoding Objectives} 
\label{sec:extrinsic_metrics_eval}

We also incorporate \metricname~into training as an alternative reward function in minimum risk training framework \cite{kumar-byrne-2004-minimum, smith-eisner-2006-minimum} and into decoding as a reranking objective in minimum Bayes risk decoding \cite{fernandes-etal-2022-quality, freitag-etal-2022-high} to improve the quality of generated simplifications.

\noindent \paragraph{Minimum Risk Training (MRT).}  Given the input $c$ and reference $r$, we generate a set of candidates $S$ for each training step and calculate the expected risk ($L_{risk}$) as follows:
\begin{equation}
   L_{risk}  = \sum_{s \in S} cost(c, r, s) \frac{P(s | c)}{\sum_{s' \in S} P(s' | c) } 
\end{equation}
\begin{equation*}
\begin{split}
  & P(s|c)  = \prod_{t=1}^{T} P(s_t | c, s_{<t};\theta) \\
  & cost(c, r, s) = 1 - Metric(c, r, s) \\
\end{split}
\end{equation*}
\noindent where $Metric(c, r, s)$ can be any evaluation metric, $\theta$ are the model parameters, and $s_{<t} = s_1, \ldots s_{t-1}$ is a partial generation of $s$.

\setlength{\tabcolsep}{5pt}
\begin{table}[t!]
\small
\centering
\begin{tabular}{lcccccc}
\toprule
& \textbf{BL}  & \textbf{SARI} & \textbf{BS} & \textbf{\metricname} &  \textbf{sBL} $\downarrow$ & \textbf{Human} \\
\midrule
MLE & 44.4 & 47.5 & \textbf{94.7} & 61.0 & 70.7 &  68.71 \\
\midrule
\multicolumn{6}{l}{\textit{Minimum Risk Training (MRT)}} \\
BL & \textbf{45.1} & 47.3 & \textbf{94.7} & 60.5 & 71.3 &  68.33 \\
SARI  & 43.4 & 48.5 & 94.5 & 63.2 & 68.1 & \underline{70.86} \\
BS & 44.0 & 48.4 & \underline{94.6} & 62.1 & 67.8 & 70.32 \\
\metricname & 42.6 & 47.4 & 94.5 & \underline{63.9} & \underline{67.1} & 68.89\\
\midrule
\multicolumn{6}{l}{\textit{Minimum Bayes Risk Decoding (MBR)}} \\
BL &  \underline{44.8} & 48.3 & 94.4 & 60.4 & 74.1 & 66.19 \\
SARI  & 43.9 & \textbf{49.6} & 94.5 & 59.3 & 75.5 & 64.31 \\
BS & 44.3 & 48.3 & \underline{94.6} & 61.1 & 73.1 & 66.79 \\
\metricname & 43.2 & \underline{49.5} & \textbf{94.7} & \textbf{73.0} & \textbf{61.8} & \textbf{72.80} 
\\
\bottomrule
\end{tabular}
\setlength{\belowcaptionskip}{-10pt}
\caption{Evaluation results for minimum risk training (MRT)  and minimum Bayes risk decoding (MBR) using T5-base model on \datasetnameSecond. We report \textbf{\metricname}, BLEU (\textbf{BL}), \textbf{SARI}, BERTScore (\textbf{BS}), self-BLEU (\textbf{sBL}), and average human ratings. MBR with \metricname~shows the best human evaluation results despite making the most number of edits to the input as indicated by its lowest self-BLEU. We use beam search with beam size of 10 for MLE.}
\label{table:mrt_mbr_eval_results}
\end{table}

Following previous work \cite{shen-etal-2016-minimum, wieting-etal-2019-beyond}, we first train the seq-to-seq generation model with  maximum likelihood estimation (MLE) for a few epochs and then we change the loss to a combination of $L_{MLE}$ and expected risk:
\begin{equation}
L_{MRT} =  \gamma L_{MLE} + (1 - \gamma) L_{risk} .     
\end{equation} 
\noindent

We choose $\gamma = 0.3$ and $|S| = 10$ for our experiments.

\noindent \paragraph{Minimum Bayes Risk Decoding (MBR).} We adopt the MBR framework proposed by \citet{fernandes-etal-2022-quality}, where we first generate a set of candidates $S$ for input $c$ during inference then rerank them by comparing each candidate $s \in S$ to all the other candidates in the set:
\begin{equation}
\hat{u}_{MBR} =  \underset{s \in S}{\arg\max} \frac{1}{|S|} \sum_{s' \in S}^{} Metric(c, s', s).      
\end{equation} 
\noindent For our experiments, we generate the candidates using beam search with beam size $=|S|$. 

\setlength{\tabcolsep}{2.5pt}
\begin{table}[t!]
\small
\centering
\begin{tabular}{lccclc}
\toprule
& \textbf{BL}  & \textbf{SARI} & \textbf{BS} & \textbf{\metricname} &  \textbf{sBL} $\downarrow$ \\
\midrule
\multicolumn{5}{l}{\textit{T5-base}} \\[0.2ex]
MLE$_{b=10}$ & \textbf{44.4} & 47.5 & 94.7 & 61.0 & 70.7  \\
MLE$_{b=100}$ & 42.6 & 46.2 & 94.3 & 57.6 (\textcolor{red}{-3.4}) & 68.1  \\
MBR-\metricname$_{|S|=10}$  & 43.6 & 48.7 & 94.6 &  67.8 (\textcolor{green!60!black}{+6.8}) & 67.0 \\
MBR-\metricname$_{|S|=100}$ & 43.2 & 49.5 & 94.7 & 73.0 (\textcolor{green!60!black}{+12.1}) & 61.8 \\
\midrule
\multicolumn{5}{l}{\textit{T5-3B}} \\[0.2ex]
MLE$_{b=10}$ & 42.3 & 46.3 &  94.8 & 60.3 & 61.9  \\
MLE$_{b=100}$ & 39.7 & 42.5 &  94.0 & 53.4 (\textcolor{red}{-6.9}) & 65.4  \\
MBR-\metricname$_{|S|=10}$  & 41.3 & 46.1 & 94.7 &  66.9 (\textcolor{green!60!black}{+6.6}) & 58.8 \\
MBR-\metricname$_{|S|=100}$  & 42.3 & \textbf{47.7} & \textbf{94.9}  & 72.6 (\textcolor{green!60!black}{+12.3}) & 55.7 \\
\midrule
\multicolumn{5}{l}{\textit{T5-11B}} \\[0.2ex]
MLE$_{b=10}$ & 37.6 & 46.4 & 93.8 & 62.9 & 49.3  \\
MLE$_{b=100}$ & 35.8 & 44.7 & 93.2 & 59.2 (\textcolor{red}{-3.7}) & 51.1  \\
MBR-\metricname$_{|S|=10}$  & 36.9 & 46.5 & 93.9 &  69.9 (\textcolor{green!60!black}{+7.0}) & 48.4 \\
MBR-\metricname$_{|S|=100}$  & 36.2 & 46.1 & 93.8 &  \textbf{74.4} (\textcolor{green!60!black}{+11.5}) & \textbf{44.6} \\
\bottomrule
\end{tabular}
\caption{Automatic evaluation results for minimum Bayes risk decoding (MBR) with different model sizes on \datasetnameSecond. For standard MLE, we use beam search with beam sizes of 10 and 100.}
\label{table:mrt_mbr_eval_results_2}
\end{table}

\setlength{\tabcolsep}{2pt}
\begin{table}[t]
\small
\centering
\resizebox{\linewidth}{!}{
\begin{tabular}{lcccccc}
\toprule
& \textbf{BL} & \textbf{SARI} & \textbf{BS}& \textbf{\metricname} &  \textbf{sBL}$\downarrow$ & \textbf{Human} \\
\midrule
\multicolumn{5}{l}{\textit{T5-11B}} \\[0.3ex]
MLE$_{b=10}$ & \textbf{37.6} & \textbf{46.4} & 93.8 & 62.9 & 49.3 & 88.80 \\
MBR-\metricname$_{|S|=100}$  & 36.2 & 46.1 & 93.8 &  \textbf{74.4} & 44.6 & 90.13  \\
\midrule
\multicolumn{5}{l}{\textit{Close-source LLMs}} \\[0.3ex]
GPT-3.5 (0-shot) & 27.8 & 41.4 & 93.4 & 60.7 & 31.8 & 90.77 \\
GPT-3.5 (5-shot) & 30.5 & 42.4 & 94.1 & 69.0 & 33.2 & 92.70 \\
GPT-4 (0-shot) & 31.6 & 43.7 & \textbf{94.3} & 73.5 & \textbf{29.1} & \textbf{93.63} \\
\bottomrule
\end{tabular}}
\setlength{\belowcaptionskip}{-10pt}
\caption{Human evaluation results for the T5-11B and close-sourced LLMs on \datasetnameSecond. T5-11B with MBR-\metricname~decoding achieves the state-of-the-art open-source model performance, on par with GPT-3.5.}
\label{table:11b_llm_human_eval}
\end{table}

\paragraph{Experiment Setup.} We fine-tune a T5 model that prepends control tokens to the input \cite{sheang-saggion-2021-controllable} to control various aspects of the generated simplification and has shown state-of-the-art performance for the task. We use \textsc{WikiAuto} \cite{jiang-etal-2020-neural} for training, \textsc{ASSET} for validation, and the challenging, complex sentences in \datasetnameSecond~ for testing. \textsc{WikiAuto} consists of 400k complex-simple sentence pairs extracted from Normal and Simple Wikipedia document pairs. For our experiments, we trained T5 towards \metricname, BLEU, SARI, and BERTScore. We provide more implementation details in Appendix \ref{app:mrt_implementation}. We report the same metrics on the test set along with self-BLEU to capture the diversity of the outputs. We also ask 3 annotators to rate the system outputs from \datasetnameSecond~ using our evaluation interface (\S \ref{sec:evaluation_framework}) and report the averaged ratings.

\paragraph{Results.} Table \ref{table:mrt_mbr_eval_results} shows that \metricname~integrated into minimal Bayes risk decoding (MBR-\metricname) achieves the best human evaluation results while it makes the most number of edits to the input (less copying) as indicated by its lowest self-BLEU score. Although MRT-\metricname~has slightly lower average human ratings than MRT-BERTScore and MRT-SARI, the pairwise comparison shows that they are very comparable: MRT-LENS generates 28\% better, 28\% worse, and 44\% equal quality simplifications compared to MRT-SARI. Additionally, BERTScore and SARI show a decrease in quality when used in decoding than standard maximum likelihood estimation (MLE).
It is noteworthy that we use a beam size of 10 for MLE rather than a large search space of beam size 100 because generation quality degrades with increased beam size as shown in Table \ref{table:mrt_mbr_eval_results_2} as well as in existing literature \cite{stahlberg-byrne-2019-nmt,meister-etal-2020-beam}.

Given the success of using \metricname~as utility function of MBR decoding on T5-base, we further apply it to larger models, including T5-3B and T5-11B. As displayed in Table \ref{table:mrt_mbr_eval_results_2}, MBR-LENS with 100 candidates improves over standard beam search by an increase of over 11 points of \metricname~score across all model sizes. Although MBR may inflate the results of the utility function it uses \cite{fernandes-etal-2022-quality}, our human evaluations solidify the assertion that T5-11B with MBR-LENS decoding exceeds standard beam search, thereby establishing state-of-the-art (SOTA) performance among open-source models. When compared to close-source large language models, T5-11B with MBR-LENS achieves on-par performance with GPT-3.5.

\setlength{\tabcolsep}{5pt}
\begin{table}[t!]
\small
\centering
\begin{tabular}{lcccc}
\toprule 
 & \textbf{\#Param} & $\mathbf{\mathlarge{\tau}_{para}}$ & 
$\mathbf{\mathlarge{\tau}_{spl}}$ & $\mathbf{\mathlarge{\tau}_{all}}$ \\
\midrule
MiniLM & 66M & 0.143 & 0.310 & 0.277 \\
BERT-base  & 110M &  0.461 & 0.240 & 0.258 \\
BERT-large & 340M & 0.461 & 0.279 & 0.298 \\
RoBERTa-base & 110M & \textbf{0.472} & 0.271 & 0.295 \\
RoBERTa-large & 340M & 0.429 & \textbf{0.333} & \textbf{0.331} \\

\bottomrule
\end{tabular}
\caption{Kendall Tau-like correlation ($\tau_{para}$ , $\tau_{spl}$, and $\tau_{all}$) of \metricname\;metric, when based on different encoder models, with human ratings in \datasetnameSecond~.}
\label{table:metric_eval_results_simpeval_model_sizes}
\end{table}
\begin{figure}[t!]
  \centering
  \includegraphics[width=0.83\linewidth]{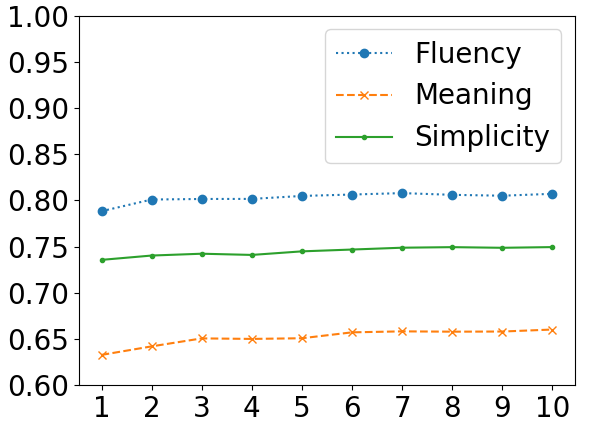}
  \caption{Pearson correlation of \metricname\;on \textsc{WikiDA} dataset using a varied number of references.}
  \label{fig:num_references}
    \vspace{-10pt}
\end{figure}

\section{LENS Analysis}
\label{sec:lens_analysis}
In this section, we delve into the impact of the underlying model architecture and the number of references on the performance of \metricname.

\paragraph{Model Architecture.} Table \ref{table:metric_eval_results_simpeval_model_sizes} shows the Kendall Tau-like correlation ($\tau_{para}$, $\tau_{spl}$, and $\tau_{all}$) of \metricname\;metric trained on various encoders on \datasetnameSecond. \metricname\;metrics trained on RoBERTa encoders \cite{Liu2019RoBERTaAR} perform better than their respective  \metricname\;metrics trained on the BERT encoders \cite{devlin-etal-2019-bert}. Among all models, \metricname~trained on RoBERTa-large achieves the highest overall correlation. It has a substantial improvement in $\tau_{spl}$ with a trade-off in $\tau_{para}$, in comparison to RoBERTa-base.

\paragraph{Number of References.} Figure \ref{fig:num_references} shows the Pearson correlation of \metricname\;metric on \textsc{Wiki-DA} for a varied number of references used during inference. Although the metric performs the best with 10 references, we see only a slight drop with one reference, demonstrating that \metricname\;is capable of evaluating with single or multiple references.

\section{\frameworkname~ Framework}
\label{sec:evaluation_framework}

\begin{figure*}[t!]

\begin{center}
\includegraphics[width=0.99\linewidth]{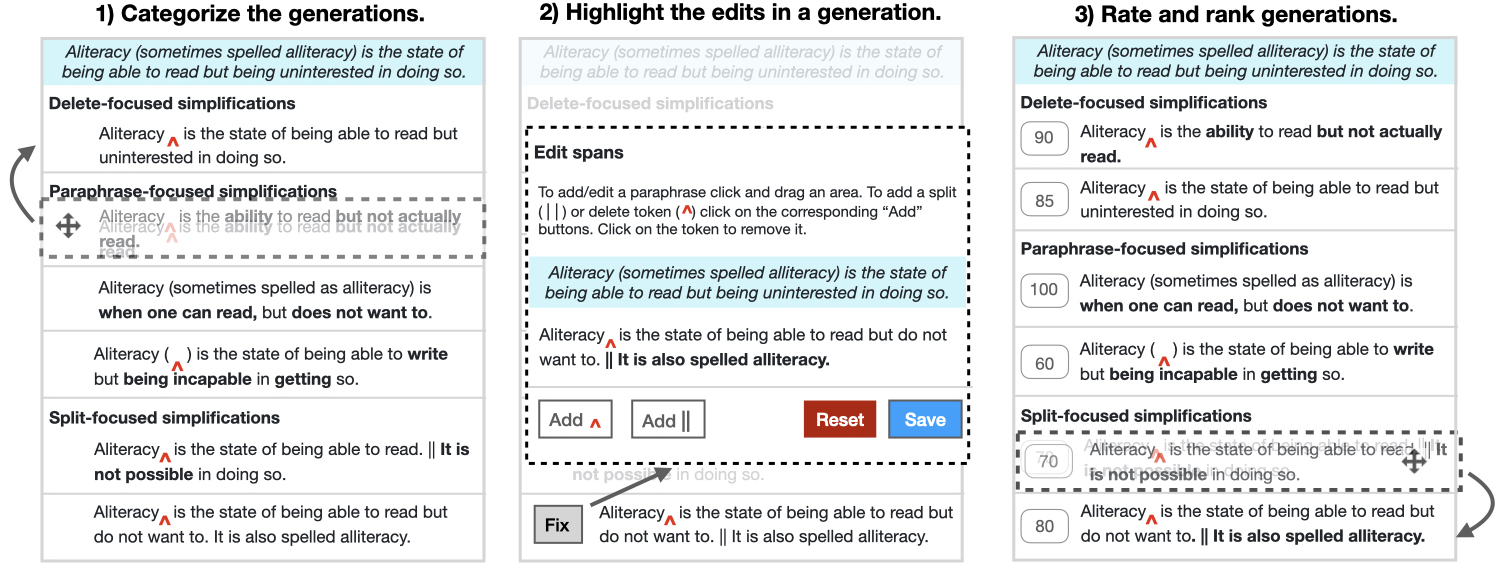}
\end{center}
\vspace{-6pt}
\caption{\frameworkname~framework consists of three steps: (1) classifying the generations, (2) annotating the edits performed by the system, and (3) rating and ranking the generations. For the first two steps, the annotators verify the automatically extracted categories and edits instead of annotating from scratch.}
\vspace{-8pt}
\label{fig:ann_framework}
\end{figure*}

To facilitate consistent evaluation and enable the training of \metricname, we develop \frameworkname~, a human evaluation framework to assist annotators in efficiently comparing and rating many ($>$20) system outputs at once in a list-wise ranking manner.

\subsection{Methodology}
We describe the three-step annotation methodology for \frameworkname~(Figure \ref{fig:ann_framework}) as follows:

\vspace{-.1cm}
\noindent \paragraph{Step 1 - Categorizing System Outputs.}
As there are different acceptable ways to simplify the input text, we first display the system outputs in groups as split-, deletion-, or paraphrase-focused based on the following criteria: (i) outputs with multiple sentences are split-focused, (ii) outputs with a compression ratio less than $0.5$ or generated by deleting words from the original sentence are deletion-focused, (iii) the rest are paraphrase-focused. Annotators can adjust the initial automatic categorization during the annotation process.

\vspace{-.1cm}
\noindent \paragraph{Step 2 - Highlighting Edits.} To help annotators notice the changes made by each system and subsequently improve the accuracy of their ratings, we use a state-of-the-art word alignment model by \citet{DBLP:journals/corr/abs-2106-02569} to highlight the edits performed by the system and classify them into three types: (i) Deletion-edits are phrases or words with no match in the modified output, (ii) Paraphrase-edits are new phrases or words added or modified, and (iii) Split-edits are any added periods (``.''). Deletion-edits are marked with a red caret (``$\wedge$''), paraphrase-edits with bolded text, and split-edits with two vertical bars (`` \textbf{||} '') (see Figure \ref{fig:ann_framework}). We ask annotators to correct the misclassified edits using an interactive pop-up window (see Appendix \ref{sec:interface}).

\vspace{-.1cm}
\noindent \paragraph{Step 3 - Ranking and Rating System Outputs.} Following the machine translation evaluation at WMT \cite{ma-etal-2018-results, ma-etal-2019-results, barrault-etal-2020-findings, akhbardeh-etal-2021-findings}, we ask annotators to rate the quality of system outputs on a 0-100 continuous scale instead of the 5-point Likert scale because the former was shown to provide higher levels of inter-annotator consistency \cite{novikova-etal-2018-rankme} and is more suitable to apply on many statistical models \cite{graham-etal-2013-continuous}. Our interactive interface allows the annotators to move the system outputs up and down to rank them and compare similar quality outputs more easily by placing them together.

\vspace{0.1cm}
\noindent We provide the annotation instructions and screenshots of the interface in Appendix \ref{sec:interface}. 

\begin{table*}[!htbp]
\small
\centering
\setlength{\tabcolsep}{3.3pt}
\begin{tabular}{lccccccccc}\toprule
 & \multicolumn{4}{c}{\textbf{\likertdatasetname}} & \multicolumn{4}{c}{\textbf{\dadatasetname}}  & \textbf{\datasetnameSecond}\\
 \cmidrule(lr){2-5} \cmidrule(lr){6-9} \cmidrule(lr){10-10}
 & \textbf{Fluency} & \textbf{Adequacy} & \textbf{Simplicity} & \textbf{Avg.} & \textbf{Fluency} & \textbf{Adequacy} & \textbf{Simplicity} & \textbf{Avg.} & \textbf{Overall} \\ \midrule
Human-1 & 4.69 & 4.46 & 1.36 & 3.50 & 92.14 & \textbf{88.87} & 83.91 & 88.30 & 81.57 \\ 
Human-2 & \textbf{4.72} & 4.48 & \textbf{1.39} & \textbf{3.53}  & 92.84 & 88.23 & \textbf{84.50} & 88.53  & \textbf{81.87}\\
GPT-3.5 (5-shot) & 4.64 & \textbf{4.66} & 1.13 & 3.48 & 92.59 & 92.10 & 81.17 & \textbf{88.62} & 79.59 \\
GPT-3.5 (zero-shot)& 4.63 & 4.56 & 0.91 & 3.37  & 90.51 & 90.32 & 74.81 & 85.21 & 75.27 \\
MUSS & 4.40 & 4.11  & 0.86 & 3.12 & 87.99 & 80.57 & 73.12 & 80.56  & 70.74 \\ 
T5-3B & 4.70 & 4.33 & 0.55 & 3.19 & \textbf{93.96} & 85.22 & 58.44 & 79.21  & 64.98 \\
\bottomrule
\end{tabular}
\caption{Model and human simplification quality under different human evaluation methods. Following \citet{sulem-etal-2018-simple}, \likertdatasetname~uses a 1 to 5 scale for fluency and adequacy, and -2 to 2 for simplicity. \dadatasetname~rates on a continuous 0-100 scale. All three methods show similar rankings of systems. \textbf{Bold}: the best.} 
\label{tab:simpeval_2022_raw}
\vspace{-8pt}
\end{table*}

\subsection{Human Evaluation Comparison}
\label{subsec:human_eval_comparison}

We compare \frameworkname~with the existing human evaluation methods: 5-point Likert \cite{sulem-etal-2018-simple} and Direct Assessment with a continuous scale of 0 to 100 \cite{alva-manchego-etal-2021-un}, which were both conducted on three dimensions: fluency, adequacy, and simplicity. For a fair comparison, we annotate the same set of simplifications using each method, resulting in \likertdatasetname~and \dadatasetname. Table \ref{tab:simpeval_2022_raw} shows similar rankings of the systems by the three methods with very slight differences. We also calculate the interval Krippendorff's $\alpha$ \cite{Krippendorff2011ComputingKA} for inter-annotator agreement. \frameworkname~achieves an $\alpha$ of 0.32, which is higher than the 0.23 $\alpha$ by Likert and 0.25 $\alpha$ by DA.
All values are considered fair \cite{krippendorff2004reliability}.

\section{Other Related Work}

\vspace{-.2cm}
 \noindent \paragraph{Text-to-text Generation Metrics.} There is currently no single automatic metric to evaluate all the text-to-text generation tasks that revise text. SARI \cite{xu-etal-2016-optimizing} is  the main metric for text simplification and has been used by other generation tasks such as sentence splitting and fusion \cite{rothe-etal-2020-leveraging, kim-etal-2021-bisect}, decontextualization \cite{DBLP:journals/corr/abs-2102-05169}, and scientific rewriting \cite{du-etal-2022-understanding-iterative}. Style transfer tasks \cite{pmlr-v70-hu17e, rao-tetreault-2018-dear, prabhumoye-etal-2018-style, krishna-etal-2020-reformulating, xu-etal-2012-paraphrasing, ma-etal-2020-powertransformer} use different automatic metrics to measure each aspect of text: (i) text similarity metrics such as BLEU, METEOR \cite{lavie-agarwal-2007-meteor}, and BERTScore to measure content preservation, (ii) text classification models \cite{sennrich-etal-2016-controlling, luo-etal-2019-towards, krishna-etal-2022-shot} or embedding-based edit distance metrics such as MoverScore \cite{zhao-etal-2019-moverscore, mir-etal-2019-evaluating} to evaluate target style, and  (iii) perplexity or pseudo log likelihood \cite{salazar-etal-2020-masked} to measure fluency. 

\vspace{-.1cm}
\noindent \paragraph{Incorporating Evaluation Metrics into Training and Inference.} Prior studies have improved MLE-trained generation systems with alternative training approaches that integrate evaluation metrics based on reinforcement learning \cite{rl-ranzato, li-etal-2016-deep, gong-etal-2019-reinforcement}, minimum risk \cite{smith-eisner-2006-minimum, shen-etal-2016-minimum}, and ranking \cite{hopkins-may-2011-tuning, xu-etal-2016-optimizing, rankgen-2022}. Incorporating metrics into decoding has also been explored by reranking the generations using discriminative rankers \cite{shen-etal-2004-discriminative, lee-etal-2021-discriminative}, energy-based rankers \cite{bhattacharyya-etal-2021-energy}, and minimum risk \cite{kumar-byrne-2004-minimum, freitag-etal-2022-high}. We use minimum risk as it has been shown to help machine translation systems \cite{wieting-etal-2019-beyond, fernandes-etal-2022-quality, amrhein-sennrich-2022-identifying}.

\section{Conclusion}
We introduce \metricname, the first supervised automatic metric for text simplification. We show that \metricname~exhibits higher human correlation than other automatic metrics. We also introduce \frameworkname~framework, which allows annotators to evaluate multiple systems' outputs at once in a list-wise manner. Using it, we create \datasetnameFirst~to train \metricname, and \datasetnameSecond~as a new metric evaluation benchmark for text simplification. We hope our metric, data, and framework will facilitate future research in text simplification evaluation. 

\section*{Limitations}
In this paper, we show that \metricname~shows better human correlation than other metrics on Wikipedia and news domains. Future research can further experiment and extend \metricname~to other domains, such as medical and children's books, as the preference for different simplification operations can vary depending on the domain and user. Additionally, our work focuses on sentence-level simplification, and future work can extend \metricname~to evaluating paragraph- and document-level simplification. \datasetname~dataset and \metricname~are also limited to the English language. 

\section*{Ethics Statement}

We used in-house annotators to collect human ratings in \datasetname~datasets and write simplifications in \datasetnameSecond. The annotators are university-level undergraduate and graduate students, including both native and non-native speakers of English. We did not collect any personal information from the annotators. We paid each annotator \$15 per hour, which is above the US federal minimum wage. We ensured that the content shown to the annotators was not upsetting and let them know that they could skip the task if they felt uncomfortable at any point. We also let the annotators know the purpose of the collected data. 

\vspace{0.1cm}
\noindent The original complex sentences in the \datasetname~datasets are from the publicly available Wikipedia. The simplifications are either from the existing work or human simplifications collected from our annotators. We used the author-released simplification outputs if they are available. For T5 (base, large, and 3B) systems, we trained our own simplification models using open-sourced code from the Hugging Face Transformers\footnote{\url{https://huggingface.co/}} library. 

\section*{Acknowledgments}
We thank Nghia T. Le, Tarek Naous, Yang Chen, and Chao Jiang as well as three anonymous reviewers for their helpful feedback. Our appreciation also extends to Colin Cherry and Alon Lavie for their valuable comments. We additionally thank Marcus Ma, Rachel Choi, Vishnesh J. Ramanathan, Elizabeth Liu, Alex Soong, Govind Ramesh, Ayush Panda, Anton Lavrouk, Vinayak Athavale, and Kelly Smith for their help with human evaluation. This research is supported in part by the NSF awards IIS-2144493 and IIS-2112633, ODNI and IARPA via the BETTER program (contract 2019-19051600004) and the HIATUS program (contract 2022-22072200004). The views and conclusions contained herein are those of the authors and should not be interpreted as necessarily representing the official policies, either expressed or implied, of NSF, ODNI, IARPA, or the U.S. Government. The U.S. Government is authorized to reproduce and distribute reprints for governmental purposes notwithstanding any copyright annotation therein.

\bibliography{anthology,custom}

\begin{thebibliography}{81}
\expandafter\ifx\csname natexlab\endcsname\relax\def\natexlab#1{#1}\fi

\bibitem[{Akhbardeh et~al.(2021)Akhbardeh, Arkhangorodsky, Biesialska, Bojar,
  Chatterjee, Chaudhary, Costa-jussa, Espa{\~n}a-Bonet, Fan, Federmann,
  Freitag, Graham, Grundkiewicz, Haddow, Harter, Heafield, Homan, Huck,
  Amponsah-Kaakyire, Kasai, Khashabi, Knight, Kocmi, Koehn, Lourie, Monz,
  Morishita, Nagata, Nagesh, Nakazawa, Negri, Pal, Tapo, Turchi, Vydrin, and
  Zampieri}]{akhbardeh-etal-2021-findings}
Farhad Akhbardeh, Arkady Arkhangorodsky, Magdalena Biesialska, Ond{\v{r}}ej
  Bojar, Rajen Chatterjee, Vishrav Chaudhary, Marta~R. Costa-jussa, Cristina
  Espa{\~n}a-Bonet, Angela Fan, Christian Federmann, Markus Freitag, Yvette
  Graham, Roman Grundkiewicz, Barry Haddow, Leonie Harter, Kenneth Heafield,
  Christopher Homan, Matthias Huck, Kwabena Amponsah-Kaakyire, Jungo Kasai,
  Daniel Khashabi, Kevin Knight, Tom Kocmi, Philipp Koehn, Nicholas Lourie,
  Christof Monz, Makoto Morishita, Masaaki Nagata, Ajay Nagesh, Toshiaki
  Nakazawa, Matteo Negri, Santanu Pal, Allahsera~Auguste Tapo, Marco Turchi,
  Valentin Vydrin, and Marcos Zampieri. 2021.
\newblock \href {https://aclanthology.org/2021.wmt-1.1} {Findings of the 2021
  conference on machine translation ({WMT}21)}.
\newblock In \emph{Proceedings of the Sixth Conference on Machine Translation},
  pages 1--88, Online. Association for Computational Linguistics.

\bibitem[{Alva-Manchego et~al.(2020)Alva-Manchego, Martin, Bordes, Scarton,
  Sagot, and Specia}]{fern2020asset}
Fernando Alva-Manchego, Louis Martin, Antoine Bordes, Carolina Scarton, Benoît
  Sagot, and Lucia Specia. 2020.
\newblock A{SSET}: {A} {D}ataset for {T}uning and {E}valuation of {S}entence
  {S}implification {M}odels with {M}ultiple {R}ewriting {T}ransformations.
\newblock In \emph{Proceedings of the Association for Computational
  Linguistics}.

\bibitem[{Alva-Manchego et~al.(2021)Alva-Manchego, Scarton, and
  Specia}]{alva-manchego-etal-2021-un}
Fernando Alva-Manchego, Carolina Scarton, and Lucia Specia. 2021.
\newblock \href {https://doi.org/10.1162/coli_a_00418} {The (un)suitability of
  automatic evaluation metrics for text simplification}.
\newblock \emph{Computational Linguistics}, 47(4):861--889.

\bibitem[{Amrhein and Sennrich(2022)}]{amrhein-sennrich-2022-identifying}
Chantal Amrhein and Rico Sennrich. 2022.
\newblock Identifying weaknesses in machine translation metrics through minimum
  {B}ayes risk decoding: A case study for {COMET}.
\newblock In \emph{Proceedings of the 2nd Conference of the Asia-Pacific
  Chapter of the Association for Computational Linguistics and the 12th
  International Joint Conference on Natural Language Processing (Volume 1: Long
  Papers)}.

\bibitem[{Artetxe et~al.(2018)Artetxe, Labaka, Agirre, and
  Cho}]{Artetxe2018UnsupervisedNM}
Mikel Artetxe, Gorka Labaka, Eneko Agirre, and Kyunghyun Cho. 2018.
\newblock Unsupervised neural machine translation.
\newblock \emph{ArXiv}, abs/1710.11041.

\bibitem[{Barrault et~al.(2020)Barrault, Biesialska, Bojar, Costa-juss{\`a},
  Federmann, Graham, Grundkiewicz, Haddow, Huck, Joanis, Kocmi, Koehn, Lo,
  Ljube{\v{s}}i{\'c}, Monz, Morishita, Nagata, Nakazawa, Pal, Post, and
  Zampieri}]{barrault-etal-2020-findings}
Lo{\"\i}c Barrault, Magdalena Biesialska, Ond{\v{r}}ej Bojar, Marta~R.
  Costa-juss{\`a}, Christian Federmann, Yvette Graham, Roman Grundkiewicz,
  Barry Haddow, Matthias Huck, Eric Joanis, Tom Kocmi, Philipp Koehn, Chi-kiu
  Lo, Nikola Ljube{\v{s}}i{\'c}, Christof Monz, Makoto Morishita, Masaaki
  Nagata, Toshiaki Nakazawa, Santanu Pal, Matt Post, and Marcos Zampieri. 2020.
\newblock \href {https://aclanthology.org/2020.wmt-1.1} {Findings of the 2020
  conference on machine translation ({WMT}20)}.
\newblock In \emph{Proceedings of the Fifth Conference on Machine Translation},
  pages 1--55, Online. Association for Computational Linguistics.

\bibitem[{Bhattacharyya et~al.(2021)Bhattacharyya, Rooshenas, Naskar, Sun,
  Iyyer, and McCallum}]{bhattacharyya-etal-2021-energy}
Sumanta Bhattacharyya, Amirmohammad Rooshenas, Subhajit Naskar, Simeng Sun,
  Mohit Iyyer, and Andrew McCallum. 2021.
\newblock \href {https://doi.org/10.18653/v1/2021.acl-long.349} {Energy-based
  reranking: Improving neural machine translation using energy-based models}.
\newblock In \emph{Proceedings of the 59th Annual Meeting of the Association
  for Computational Linguistics and the 11th International Joint Conference on
  Natural Language Processing (Volume 1: Long Papers)}, pages 4528--4537,
  Online. Association for Computational Linguistics.

\bibitem[{Bojar et~al.(2017)Bojar, Graham, and
  Kamran}]{bojar-etal-2017-results}
Ond{\v{r}}ej Bojar, Yvette Graham, and Amir Kamran. 2017.
\newblock \href {https://doi.org/10.18653/v1/W17-4755} {Results of the {WMT}17
  metrics shared task}.
\newblock In \emph{Proceedings of the Second Conference on Machine
  Translation}, pages 489--513, Copenhagen, Denmark. Association for
  Computational Linguistics.

\bibitem[{Choi et~al.(2021)Choi, Palomaki, Lamm, Kwiatkowski, Das, and
  Collins}]{DBLP:journals/corr/abs-2102-05169}
Eunsol Choi, Jennimaria Palomaki, Matthew Lamm, Tom Kwiatkowski, Dipanjan Das,
  and Michael Collins. 2021.
\newblock Decontextualization: Making sentences stand-alone.
\newblock \emph{Transactions of the Association for Computational Linguistics
  (TACL)}.

\bibitem[{Chung et~al.(2022)Chung, Hou, Longpre, Zoph, Tay, Fedus, Li, Wang,
  Dehghani, Brahma et~al.}]{chung2022scaling}
Hyung~Won Chung, Le~Hou, Shayne Longpre, Barret Zoph, Yi~Tay, William Fedus,
  Eric Li, Xuezhi Wang, Mostafa Dehghani, Siddhartha Brahma, et~al. 2022.
\newblock Scaling instruction-finetuned language models.
\newblock \emph{arXiv preprint arXiv:2210.11416}.

\bibitem[{Deutsch et~al.(2021)Deutsch, Dror, and
  Roth}]{deutsch-etal-2021-statistical}
Daniel Deutsch, Rotem Dror, and Dan Roth. 2021.
\newblock \href {https://doi.org/10.1162/tacl_a_00417} {A statistical analysis
  of summarization evaluation metrics using resampling methods}.
\newblock \emph{Transactions of the Association for Computational Linguistics},
  9:1132--1146.

\bibitem[{Devlin et~al.(2019)Devlin, Chang, Lee, and
  Toutanova}]{devlin-etal-2019-bert}
Jacob Devlin, Ming-Wei Chang, Kenton Lee, and Kristina Toutanova. 2019.
\newblock \href {https://doi.org/10.18653/v1/N19-1423} {{BERT}: Pre-training of
  deep bidirectional transformers for language understanding}.
\newblock In \emph{Proceedings of the 2019 Conference of the North {A}merican
  Chapter of the Association for Computational Linguistics: Human Language
  Technologies, Volume 1 (Long and Short Papers)}, pages 4171--4186,
  Minneapolis, Minnesota. Association for Computational Linguistics.

\bibitem[{Dong et~al.(2019)Dong, Li, Rezagholizadeh, and
  Cheung}]{dong-etal-2019-editnts}
Yue Dong, Zichao Li, Mehdi Rezagholizadeh, and Jackie Chi~Kit Cheung. 2019.
\newblock \href {https://doi.org/10.18653/v1/P19-1331} {{E}dit{NTS}: An neural
  programmer-interpreter model for sentence simplification through explicit
  editing}.
\newblock In \emph{Proceedings of the 57th Annual Meeting of the Association
  for Computational Linguistics}, pages 3393--3402, Florence, Italy.
  Association for Computational Linguistics.

\bibitem[{Dou et~al.(2022)Dou, Forbes, Koncel-Kedziorski, Smith, and
  Choi}]{dou-etal-2022-gpt}
Yao Dou, Maxwell Forbes, Rik Koncel-Kedziorski, Noah~A. Smith, and Yejin Choi.
  2022.
\newblock \href {https://doi.org/10.18653/v1/2022.acl-long.501} {Is {GPT}-3
  text indistinguishable from human text? scarecrow: A framework for
  scrutinizing machine text}.
\newblock In \emph{Proceedings of the 60th Annual Meeting of the Association
  for Computational Linguistics (Volume 1: Long Papers)}, pages 7250--7274,
  Dublin, Ireland. Association for Computational Linguistics.

\bibitem[{Du et~al.(2022)Du, Raheja, Kumar, Kim, Lopez, and
  Kang}]{du-etal-2022-understanding-iterative}
Wanyu Du, Vipul Raheja, Dhruv Kumar, Zae~Myung Kim, Melissa Lopez, and Dongyeop
  Kang. 2022.
\newblock \href {https://doi.org/10.18653/v1/2022.acl-long.250} {Understanding
  iterative revision from human-written text}.
\newblock In \emph{Proceedings of the 60th Annual Meeting of the Association
  for Computational Linguistics (Volume 1: Long Papers)}, pages 3573--3590,
  Dublin, Ireland. Association for Computational Linguistics.

\bibitem[{Fernandes et~al.(2022)Fernandes, Farinhas, Rei, De~Souza, Ogayo,
  Neubig, and Martins}]{fernandes-etal-2022-quality}
Patrick Fernandes, Ant{\'o}nio Farinhas, Ricardo Rei, Jos{\'e} De~Souza, Perez
  Ogayo, Graham Neubig, and Andre Martins. 2022.
\newblock \href {https://doi.org/10.18653/v1/2022.naacl-main.100}
  {Quality-aware decoding for neural machine translation}.
\newblock In \emph{Proceedings of the 2022 Conference of the North American
  Chapter of the Association for Computational Linguistics: Human Language
  Technologies}, pages 1396--1412, Seattle, United States. Association for
  Computational Linguistics.

\bibitem[{Freitag et~al.(2022)Freitag, Grangier, Tan, and
  Liang}]{freitag-etal-2022-high}
Markus Freitag, David Grangier, Qijun Tan, and Bowen Liang. 2022.
\newblock \href {https://doi.org/10.1162/tacl_a_00491} {High quality rather
  than high model probability: Minimum {B}ayes risk decoding with neural
  metrics}.
\newblock \emph{Transactions of the Association for Computational Linguistics},
  10:811--825.

\bibitem[{Gong et~al.(2019)Gong, Bhat, Wu, Xiong, and
  Hwu}]{gong-etal-2019-reinforcement}
Hongyu Gong, Suma Bhat, Lingfei Wu, JinJun Xiong, and Wen-mei Hwu. 2019.
\newblock \href {https://doi.org/10.18653/v1/N19-1320} {Reinforcement learning
  based text style transfer without parallel training corpus}.
\newblock In \emph{Proceedings of the 2019 Conference of the North {A}merican
  Chapter of the Association for Computational Linguistics: Human Language
  Technologies, Volume 1 (Long and Short Papers)}, pages 3168--3180,
  Minneapolis, Minnesota. Association for Computational Linguistics.

\bibitem[{Goyal et~al.(2022)Goyal, Li, and Durrett}]{goyal2022news}
Tanya Goyal, Junyi~Jessy Li, and Greg Durrett. 2022.
\newblock News summarization and evaluation in the era of {GPT}-3.
\newblock \emph{arXiv preprint arXiv:2209.12356}.

\bibitem[{Graham et~al.(2013)Graham, Baldwin, Moffat, and
  Zobel}]{graham-etal-2013-continuous}
Yvette Graham, Timothy Baldwin, Alistair Moffat, and Justin Zobel. 2013.
\newblock \href {https://aclanthology.org/W13-2305} {Continuous measurement
  scales in human evaluation of machine translation}.
\newblock In \emph{Proceedings of the 7th Linguistic Annotation Workshop and
  Interoperability with Discourse}, pages 33--41, Sofia, Bulgaria. Association
  for Computational Linguistics.

\bibitem[{Hopkins and May(2011)}]{hopkins-may-2011-tuning}
Mark Hopkins and Jonathan May. 2011.
\newblock \href {https://aclanthology.org/D11-1125} {Tuning as ranking}.
\newblock In \emph{Proceedings of the 2011 Conference on Empirical Methods in
  Natural Language Processing}, pages 1352--1362, Edinburgh, Scotland, UK.
  Association for Computational Linguistics.

\bibitem[{Hu et~al.(2017)Hu, Yang, Liang, Salakhutdinov, and
  Xing}]{pmlr-v70-hu17e}
Zhiting Hu, Zichao Yang, Xiaodan Liang, Ruslan Salakhutdinov, and Eric~P. Xing.
  2017.
\newblock Toward controlled generation of text.
\newblock In \emph{Proceedings of the 34th International Conference on Machine
  Learning}.

\bibitem[{Jiang et~al.(2020)Jiang, Maddela, Lan, Zhong, and
  Xu}]{jiang-etal-2020-neural}
Chao Jiang, Mounica Maddela, Wuwei Lan, Yang Zhong, and Wei Xu. 2020.
\newblock \href {https://doi.org/10.18653/v1/2020.acl-main.709} {Neural {CRF}
  model for sentence alignment in text simplification}.
\newblock In \emph{Proceedings of the 58th Annual Meeting of the Association
  for Computational Linguistics}, pages 7943--7960, Online. Association for
  Computational Linguistics.

\bibitem[{Kauchak(2013)}]{kauchak-2013-improving}
David Kauchak. 2013.
\newblock \href {https://aclanthology.org/P13-1151} {Improving text
  simplification language modeling using unsimplified text data}.
\newblock In \emph{Proceedings of the 51st Annual Meeting of the Association
  for Computational Linguistics (Volume 1: Long Papers)}, pages 1537--1546,
  Sofia, Bulgaria. Association for Computational Linguistics.

\bibitem[{Kim et~al.(2021)Kim, Maddela, Kriz, Xu, and
  Callison-Burch}]{kim-etal-2021-bisect}
Joongwon Kim, Mounica Maddela, Reno Kriz, Wei Xu, and Chris Callison-Burch.
  2021.
\newblock \href {https://doi.org/10.18653/v1/2021.emnlp-main.500} {{B}i{SECT}:
  Learning to split and rephrase sentences with bitexts}.
\newblock In \emph{Proceedings of the 2021 Conference on Empirical Methods in
  Natural Language Processing}, pages 6193--6209, Online and Punta Cana,
  Dominican Republic. Association for Computational Linguistics.

\bibitem[{Kincaid(1975)}]{kincaid}
Kincaid. 1975.
\newblock Derivation of new readability formulas (automated readability index,
  fog count and flesch reading ease formula) for navy enlisted personnel.
\newblock \emph{research branch report}.

\bibitem[{Krippendorff(2004)}]{krippendorff2004reliability}
Klaus Krippendorff. 2004.
\newblock Reliability in content analysis: Some common misconceptions and
  recommendations.
\newblock \emph{Human communication research}, 30(3):411--433.

\bibitem[{Krippendorff(2011)}]{Krippendorff2011ComputingKA}
Klaus Krippendorff. 2011.
\newblock Computing krippendorff's alpha-reliability.

\bibitem[{Krishna et~al.(2022{\natexlab{a}})Krishna, Chang, Wieting, and
  Iyyer}]{rankgen-2022}
Kalpesh Krishna, Yapei Chang, John Wieting, and Mohit Iyyer.
  2022{\natexlab{a}}.
\newblock Rankgen: Improving text generation with large ranking models.

\bibitem[{Krishna et~al.(2022{\natexlab{b}})Krishna, Nathani, Garcia, Samanta,
  and Talukdar}]{krishna-etal-2022-shot}
Kalpesh Krishna, Deepak Nathani, Xavier Garcia, Bidisha Samanta, and Partha
  Talukdar. 2022{\natexlab{b}}.
\newblock \href {https://doi.org/10.18653/v1/2022.acl-long.514} {Few-shot
  controllable style transfer for low-resource multilingual settings}.
\newblock In \emph{Proceedings of the 60th Annual Meeting of the Association
  for Computational Linguistics (Volume 1: Long Papers)}, pages 7439--7468,
  Dublin, Ireland. Association for Computational Linguistics.

\bibitem[{Krishna et~al.(2020)Krishna, Wieting, and
  Iyyer}]{krishna-etal-2020-reformulating}
Kalpesh Krishna, John Wieting, and Mohit Iyyer. 2020.
\newblock \href {https://doi.org/10.18653/v1/2020.emnlp-main.55} {Reformulating
  unsupervised style transfer as paraphrase generation}.
\newblock In \emph{Proceedings of the 2020 Conference on Empirical Methods in
  Natural Language Processing (EMNLP)}, pages 737--762, Online. Association for
  Computational Linguistics.

\bibitem[{Kumar et~al.(2020)Kumar, Mou, Golab, and Olga}]{dhruv-acl-2020}
Dhruv Kumar, Lili Mou, Lukasz Golab, and Vechtomova Olga. 2020.
\newblock Iterative edit-based unsupervised sentence simplification.
\newblock In \emph{Proceedings of the Association for Computational
  Linguistics}.

\bibitem[{Kumar and Byrne(2004)}]{kumar-byrne-2004-minimum}
Shankar Kumar and William Byrne. 2004.
\newblock \href {https://aclanthology.org/N04-1022} {Minimum {B}ayes-risk
  decoding for statistical machine translation}.
\newblock In \emph{Proceedings of the Human Language Technology Conference of
  the North {A}merican Chapter of the Association for Computational
  Linguistics: {HLT}-{NAACL} 2004}, pages 169--176, Boston, Massachusetts, USA.
  Association for Computational Linguistics.

\bibitem[{Lan et~al.(2021)Lan, Jiang, and
  Xu}]{DBLP:journals/corr/abs-2106-02569}
Wuwei Lan, Chao Jiang, and Wei Xu. 2021.
\newblock Neural semi-markov {CRF} for monolingual word alignment.

\bibitem[{Lavie and Agarwal(2007)}]{lavie-agarwal-2007-meteor}
Alon Lavie and Abhaya Agarwal. 2007.
\newblock \href {https://aclanthology.org/W07-0734} {{METEOR}: An automatic
  metric for {MT} evaluation with high levels of correlation with human
  judgments}.
\newblock In \emph{Proceedings of the Second Workshop on Statistical Machine
  Translation}, pages 228--231, Prague, Czech Republic. Association for
  Computational Linguistics.

\bibitem[{Lee et~al.(2021)Lee, Auli, and
  Ranzato}]{lee-etal-2021-discriminative}
Ann Lee, Michael Auli, and Marc{'}Aurelio Ranzato. 2021.
\newblock \href {https://doi.org/10.18653/v1/2021.acl-long.563} {Discriminative
  reranking for neural machine translation}.
\newblock In \emph{Proceedings of the 59th Annual Meeting of the Association
  for Computational Linguistics and the 11th International Joint Conference on
  Natural Language Processing (Volume 1: Long Papers)}, pages 7250--7264,
  Online. Association for Computational Linguistics.

\bibitem[{Lewis et~al.(2020)Lewis, Liu, Goyal, Ghazvininejad, Mohamed, Levy,
  Stoyanov, and Zettlemoyer}]{lewis-etal-2020-bart}
Mike Lewis, Yinhan Liu, Naman Goyal, Marjan Ghazvininejad, Abdelrahman Mohamed,
  Omer Levy, Veselin Stoyanov, and Luke Zettlemoyer. 2020.
\newblock \href {https://doi.org/10.18653/v1/2020.acl-main.703} {{BART}:
  Denoising sequence-to-sequence pre-training for natural language generation,
  translation, and comprehension}.
\newblock In \emph{Proceedings of the 58th Annual Meeting of the Association
  for Computational Linguistics}, pages 7871--7880, Online. Association for
  Computational Linguistics.

\bibitem[{Li et~al.(2016)Li, Monroe, Ritter, Jurafsky, Galley, and
  Gao}]{li-etal-2016-deep}
Jiwei Li, Will Monroe, Alan Ritter, Dan Jurafsky, Michel Galley, and Jianfeng
  Gao. 2016.
\newblock \href {https://doi.org/10.18653/v1/D16-1127} {Deep reinforcement
  learning for dialogue generation}.
\newblock In \emph{Proceedings of the 2016 Conference on Empirical Methods in
  Natural Language Processing}, pages 1192--1202, Austin, Texas. Association
  for Computational Linguistics.

\bibitem[{Liu et~al.(2019)Liu, Ott, Goyal, Du, Joshi, Chen, Levy, Lewis,
  Zettlemoyer, and Stoyanov}]{Liu2019RoBERTaAR}
Yinhan Liu, Myle Ott, Naman Goyal, Jingfei Du, Mandar Joshi, Danqi Chen, Omer
  Levy, Mike Lewis, Luke Zettlemoyer, and Veselin Stoyanov. 2019.
\newblock {RoBERTa}: A robustly optimized {BERT} pretraining approach.
\newblock \emph{ArXiv}, abs/1907.11692.

\bibitem[{Loshchilov and Hutter(2017)}]{loshchilov2017decoupled}
Ilya Loshchilov and Frank Hutter. 2017.
\newblock Decoupled weight decay regularization.
\newblock \emph{arXiv preprint arXiv:1711.05101}.

\bibitem[{Luo et~al.(2019)Luo, Li, Yang, Zhou, Tan, Chang, Sui, and
  Sun}]{luo-etal-2019-towards}
Fuli Luo, Peng Li, Pengcheng Yang, Jie Zhou, Yutong Tan, Baobao Chang, Zhifang
  Sui, and Xu~Sun. 2019.
\newblock \href {https://doi.org/10.18653/v1/P19-1194} {Towards fine-grained
  text sentiment transfer}.
\newblock In \emph{Proceedings of the 57th Annual Meeting of the Association
  for Computational Linguistics}, pages 2013--2022, Florence, Italy.
  Association for Computational Linguistics.

\bibitem[{Ma et~al.(2018)Ma, Bojar, and Graham}]{ma-etal-2018-results}
Qingsong Ma, Ond{\v{r}}ej Bojar, and Yvette Graham. 2018.
\newblock \href {https://doi.org/10.18653/v1/W18-6450} {Results of the {WMT}18
  metrics shared task: Both characters and embeddings achieve good
  performance}.
\newblock In \emph{Proceedings of the Third Conference on Machine Translation:
  Shared Task Papers}, pages 671--688, Belgium, Brussels. Association for
  Computational Linguistics.

\bibitem[{Ma et~al.(2019)Ma, Wei, Bojar, and Graham}]{ma-etal-2019-results}
Qingsong Ma, Johnny Wei, Ond{\v{r}}ej Bojar, and Yvette Graham. 2019.
\newblock \href {https://doi.org/10.18653/v1/W19-5302} {Results of the {WMT}19
  metrics shared task: Segment-level and strong {MT} systems pose big
  challenges}.
\newblock In \emph{Proceedings of the Fourth Conference on Machine Translation
  (Volume 2: Shared Task Papers, Day 1)}, pages 62--90, Florence, Italy.
  Association for Computational Linguistics.

\bibitem[{Ma et~al.(2020)Ma, Sap, Rashkin, and
  Choi}]{ma-etal-2020-powertransformer}
Xinyao Ma, Maarten Sap, Hannah Rashkin, and Yejin Choi. 2020.
\newblock \href {https://doi.org/10.18653/v1/2020.emnlp-main.602}
  {{P}ower{T}ransformer: Unsupervised controllable revision for biased language
  correction}.
\newblock In \emph{Proceedings of the 2020 Conference on Empirical Methods in
  Natural Language Processing (EMNLP)}, pages 7426--7441, Online. Association
  for Computational Linguistics.

\bibitem[{Maddela et~al.(2021)Maddela, Alva-Manchego, and
  Xu}]{maddela-etal-2021-controllable}
Mounica Maddela, Fernando Alva-Manchego, and Wei Xu. 2021.
\newblock \href {https://doi.org/10.18653/v1/2021.naacl-main.277} {Controllable
  text simplification with explicit paraphrasing}.
\newblock In \emph{Proceedings of the 2021 Conference of the North American
  Chapter of the Association for Computational Linguistics: Human Language
  Technologies}, pages 3536--3553, Online. Association for Computational
  Linguistics.

\bibitem[{Martin et~al.(2020)Martin, de~la Clergerie, Sagot, and
  Bordes}]{martin-etal-2020-controllable}
Louis Martin, {\'E}ric de~la Clergerie, Beno{\^\i}t Sagot, and Antoine Bordes.
  2020.
\newblock \href {https://aclanthology.org/2020.lrec-1.577} {Controllable
  sentence simplification}.
\newblock In \emph{Proceedings of the Twelfth Language Resources and Evaluation
  Conference}, pages 4689--4698, Marseille, France. European Language Resources
  Association.

\bibitem[{Martin et~al.(2022)Martin, Fan, de~la Clergerie, Bordes, and
  Sagot}]{martin-etal-2022-muss}
Louis Martin, Angela Fan, {\'E}ric de~la Clergerie, Antoine Bordes, and
  Beno{\^\i}t Sagot. 2022.
\newblock \href {https://aclanthology.org/2022.lrec-1.176} {{MUSS}:
  Multilingual unsupervised sentence simplification by mining paraphrases}.
\newblock In \emph{Proceedings of the Thirteenth Language Resources and
  Evaluation Conference}, pages 1651--1664, Marseille, France. European
  Language Resources Association.

\bibitem[{Meister et~al.(2020)Meister, Cotterell, and
  Vieira}]{meister-etal-2020-beam}
Clara Meister, Ryan Cotterell, and Tim Vieira. 2020.
\newblock \href {https://doi.org/10.18653/v1/2020.emnlp-main.170} {If beam
  search is the answer, what was the question?}
\newblock In \emph{Proceedings of the 2020 Conference on Empirical Methods in
  Natural Language Processing (EMNLP)}, pages 2173--2185, Online. Association
  for Computational Linguistics.

\bibitem[{Mir et~al.(2019)Mir, Felbo, Obradovich, and
  Rahwan}]{mir-etal-2019-evaluating}
Remi Mir, Bjarke Felbo, Nick Obradovich, and Iyad Rahwan. 2019.
\newblock \href {https://doi.org/10.18653/v1/N19-1049} {Evaluating style
  transfer for text}.
\newblock In \emph{Proceedings of the 2019 Conference of the North {A}merican
  Chapter of the Association for Computational Linguistics: Human Language
  Technologies, Volume 1 (Long and Short Papers)}, pages 495--504, Minneapolis,
  Minnesota. Association for Computational Linguistics.

\bibitem[{Narayan and Gardent(2014)}]{narayan-gardent-2014-hybrid}
Shashi Narayan and Claire Gardent. 2014.
\newblock \href {https://doi.org/10.3115/v1/P14-1041} {Hybrid simplification
  using deep semantics and machine translation}.
\newblock In \emph{Proceedings of the 52nd Annual Meeting of the Association
  for Computational Linguistics (Volume 1: Long Papers)}, pages 435--445,
  Baltimore, Maryland. Association for Computational Linguistics.

\bibitem[{Novikova et~al.(2018)Novikova, Du{\v{s}}ek, and
  Rieser}]{novikova-etal-2018-rankme}
Jekaterina Novikova, Ond{\v{r}}ej Du{\v{s}}ek, and Verena Rieser. 2018.
\newblock \href {https://doi.org/10.18653/v1/N18-2012} {{R}ank{ME}: Reliable
  human ratings for natural language generation}.
\newblock In \emph{Proceedings of the 2018 Conference of the North {A}merican
  Chapter of the Association for Computational Linguistics: Human Language
  Technologies, Volume 2 (Short Papers)}, pages 72--78, New Orleans, Louisiana.
  Association for Computational Linguistics.

\bibitem[{Ouyang et~al.(2022)Ouyang, Wu, Jiang, Almeida, Wainwright, Mishkin,
  Zhang, Agarwal, Slama, Ray et~al.}]{ouyang2022training}
Long Ouyang, Jeff Wu, Xu~Jiang, Diogo Almeida, Carroll~L Wainwright, Pamela
  Mishkin, Chong Zhang, Sandhini Agarwal, Katarina Slama, Alex Ray, et~al.
  2022.
\newblock Training language models to follow instructions with human feedback.
\newblock \emph{arXiv preprint arXiv:2203.02155}.

\bibitem[{Pang and Gimpel(2019)}]{pang-gimpel-2019-unsupervised}
Richard~Yuanzhe Pang and Kevin Gimpel. 2019.
\newblock \href {https://doi.org/10.18653/v1/D19-5614} {Unsupervised evaluation
  metrics and learning criteria for non-parallel textual transfer}.
\newblock In \emph{Proceedings of the 3rd Workshop on Neural Generation and
  Translation}, pages 138--147, Hong Kong. Association for Computational
  Linguistics.

\bibitem[{Prabhumoye et~al.(2018)Prabhumoye, Tsvetkov, Salakhutdinov, and
  Black}]{prabhumoye-etal-2018-style}
Shrimai Prabhumoye, Yulia Tsvetkov, Ruslan Salakhutdinov, and Alan~W Black.
  2018.
\newblock \href {https://doi.org/10.18653/v1/P18-1080} {Style transfer through
  back-translation}.
\newblock In \emph{Proceedings of the 56th Annual Meeting of the Association
  for Computational Linguistics (Volume 1: Long Papers)}, pages 866--876,
  Melbourne, Australia. Association for Computational Linguistics.

\bibitem[{Raffel et~al.(2020)Raffel, Shazeer, Roberts, Lee, Narang, Matena,
  Zhou, Li, Liu et~al.}]{raffel2020exploring}
Colin Raffel, Noam Shazeer, Adam Roberts, Katherine Lee, Sharan Narang, Michael
  Matena, Yanqi Zhou, Wei Li, Peter~J Liu, et~al. 2020.
\newblock Exploring the limits of transfer learning with a unified text-to-text
  transformer.
\newblock \emph{J. Mach. Learn. Res.}, 21(140):1--67.

\bibitem[{Ranzato et~al.(2015)Ranzato, Chopra, Auli, and Zaremba}]{rl-ranzato}
Marc'Aurelio Ranzato, Sumit Chopra, Michael Auli, and Wojciech Zaremba. 2015.
\newblock Sequence level training with recurrent neural networks.

\bibitem[{Rao and Tetreault(2018)}]{rao-tetreault-2018-dear}
Sudha Rao and Joel Tetreault. 2018.
\newblock \href {https://doi.org/10.18653/v1/N18-1012} {Dear sir or madam, may
  {I} introduce the {GYAFC} dataset: Corpus, benchmarks and metrics for
  formality style transfer}.
\newblock In \emph{Proceedings of the 2018 Conference of the North {A}merican
  Chapter of the Association for Computational Linguistics: Human Language
  Technologies, Volume 1 (Long Papers)}, pages 129--140, New Orleans,
  Louisiana. Association for Computational Linguistics.

\bibitem[{Rei et~al.(2020)Rei, Stewart, Farinha, and
  Lavie}]{rei-etal-2020-comet}
Ricardo Rei, Craig Stewart, Ana~C Farinha, and Alon Lavie. 2020.
\newblock \href {https://doi.org/10.18653/v1/2020.emnlp-main.213} {{COMET}: A
  neural framework for {MT} evaluation}.
\newblock In \emph{Proceedings of the 2020 Conference on Empirical Methods in
  Natural Language Processing (EMNLP)}, pages 2685--2702, Online. Association
  for Computational Linguistics.

\bibitem[{Rothe et~al.(2020)Rothe, Narayan, and
  Severyn}]{rothe-etal-2020-leveraging}
Sascha Rothe, Shashi Narayan, and Aliaksei Severyn. 2020.
\newblock \href {https://doi.org/10.1162/tacl_a_00313} {Leveraging pre-trained
  checkpoints for sequence generation tasks}.
\newblock \emph{Transactions of the Association for Computational Linguistics},
  8:264--280.

\bibitem[{Saggion(2017)}]{Saiggon}
Horacio Saggion. 2017.
\newblock Automatic text simplification.
\newblock \emph{Synthesis Lectures on Human Language Technologies}.

\bibitem[{Salazar et~al.(2020)Salazar, Liang, Nguyen, and
  Kirchhoff}]{salazar-etal-2020-masked}
Julian Salazar, Davis Liang, Toan~Q. Nguyen, and Katrin Kirchhoff. 2020.
\newblock \href {https://doi.org/10.18653/v1/2020.acl-main.240} {Masked
  language model scoring}.
\newblock In \emph{Proceedings of the 58th Annual Meeting of the Association
  for Computational Linguistics}, pages 2699--2712, Online. Association for
  Computational Linguistics.

\bibitem[{Sellam et~al.(2020)Sellam, Das, and Parikh}]{sellam-etal-2020-bleurt}
Thibault Sellam, Dipanjan Das, and Ankur Parikh. 2020.
\newblock \href {https://doi.org/10.18653/v1/2020.acl-main.704} {{BLEURT}:
  Learning robust metrics for text generation}.
\newblock In \emph{Proceedings of the 58th Annual Meeting of the Association
  for Computational Linguistics}, pages 7881--7892, Online. Association for
  Computational Linguistics.

\bibitem[{Sennrich et~al.(2016)Sennrich, Haddow, and
  Birch}]{sennrich-etal-2016-controlling}
Rico Sennrich, Barry Haddow, and Alexandra Birch. 2016.
\newblock \href {https://doi.org/10.18653/v1/N16-1005} {Controlling politeness
  in neural machine translation via side constraints}.
\newblock In \emph{Proceedings of the 2016 Conference of the North {A}merican
  Chapter of the Association for Computational Linguistics: Human Language
  Technologies}, pages 35--40, San Diego, California. Association for
  Computational Linguistics.

\bibitem[{Sheang and Saggion(2021)}]{sheang-saggion-2021-controllable}
Kim~Cheng Sheang and Horacio Saggion. 2021.
\newblock \href {https://aclanthology.org/2021.inlg-1.38} {Controllable
  sentence simplification with a unified text-to-text transfer transformer}.
\newblock In \emph{Proceedings of the 14th International Conference on Natural
  Language Generation}, pages 341--352, Aberdeen, Scotland, UK. Association for
  Computational Linguistics.

\bibitem[{Shen et~al.(2004)Shen, Sarkar, and
  Och}]{shen-etal-2004-discriminative}
Libin Shen, Anoop Sarkar, and Franz~Josef Och. 2004.
\newblock \href {https://aclanthology.org/N04-1023} {Discriminative reranking
  for machine translation}.
\newblock In \emph{Proceedings of the Human Language Technology Conference of
  the North {A}merican Chapter of the Association for Computational
  Linguistics: {HLT}-{NAACL} 2004}, pages 177--184, Boston, Massachusetts, USA.
  Association for Computational Linguistics.

\bibitem[{Shen et~al.(2016)Shen, Cheng, He, He, Wu, Sun, and
  Liu}]{shen-etal-2016-minimum}
Shiqi Shen, Yong Cheng, Zhongjun He, Wei He, Hua Wu, Maosong Sun, and Yang Liu.
  2016.
\newblock \href {https://doi.org/10.18653/v1/P16-1159} {Minimum risk training
  for neural machine translation}.
\newblock In \emph{Proceedings of the 54th Annual Meeting of the Association
  for Computational Linguistics (Volume 1: Long Papers)}, pages 1683--1692,
  Berlin, Germany. Association for Computational Linguistics.

\bibitem[{Smith and Eisner(2006)}]{smith-eisner-2006-minimum}
David~A. Smith and Jason Eisner. 2006.
\newblock \href {https://aclanthology.org/P06-2101} {Minimum risk annealing for
  training log-linear models}.
\newblock In \emph{Proceedings of the {COLING}/{ACL} 2006 Main Conference
  Poster Sessions}, pages 787--794, Sydney, Australia. Association for
  Computational Linguistics.

\bibitem[{Stahlberg and Byrne(2019)}]{stahlberg-byrne-2019-nmt}
Felix Stahlberg and Bill Byrne. 2019.
\newblock \href {https://doi.org/10.18653/v1/D19-1331} {On {NMT} search errors
  and model errors: Cat got your tongue?}
\newblock In \emph{Proceedings of the 2019 Conference on Empirical Methods in
  Natural Language Processing and the 9th International Joint Conference on
  Natural Language Processing (EMNLP-IJCNLP)}, pages 3356--3362, Hong Kong,
  China. Association for Computational Linguistics.

\bibitem[{Sulem(2018)}]{sulem-etal-2018-semantic-short}
Elior Sulem. 2018.
\newblock Semantic structural evaluation for text simplification.
\newblock In \emph{Proceedings of the 2018 Conference of the North {A}merican
  Chapter of the Association for Computational Linguistics: Human Language
  Technologies, Volume 1 (Long Papers)}.

\bibitem[{Sulem et~al.(2018)Sulem, Abend, and
  Rappoport}]{sulem-etal-2018-simple}
Elior Sulem, Omri Abend, and Ari Rappoport. 2018.
\newblock \href {https://doi.org/10.18653/v1/P18-1016} {Simple and effective
  text simplification using semantic and neural methods}.
\newblock In \emph{Proceedings of the 56th Annual Meeting of the Association
  for Computational Linguistics (Volume 1: Long Papers)}, pages 162--173,
  Melbourne, Australia. Association for Computational Linguistics.

\bibitem[{Surya et~al.(2019)Surya, Mishra, Laha, Jain, and
  Sankaranarayanan}]{surya-etal-2019-unsupervised}
Sai Surya, Abhijit Mishra, Anirban Laha, Parag Jain, and Karthik
  Sankaranarayanan. 2019.
\newblock \href {https://doi.org/10.18653/v1/P19-1198} {Unsupervised neural
  text simplification}.
\newblock In \emph{Proceedings of the 57th Annual Meeting of the Association
  for Computational Linguistics}, pages 2058--2068, Florence, Italy.
  Association for Computational Linguistics.

\bibitem[{Vaswani et~al.(2017)Vaswani, Shazeer, Parmar, Uszkoreit, Jones,
  Gomez, Kaiser, and Polosukhin}]{transformer-vaswani-2017}
Ashish Vaswani, Noam Shazeer, Niki Parmar, Jakob Uszkoreit, Llion Jones,
  Aidan~N. Gomez, \L{}ukasz Kaiser, and Illia Polosukhin. 2017.
\newblock Attention is all you need.
\newblock In \emph{Proceedings of the 31st International Conference on Neural
  Information Processing Systems}.

\bibitem[{Wieting et~al.(2019)Wieting, Berg-Kirkpatrick, Gimpel, and
  Neubig}]{wieting-etal-2019-beyond}
John Wieting, Taylor Berg-Kirkpatrick, Kevin Gimpel, and Graham Neubig. 2019.
\newblock \href {https://doi.org/10.18653/v1/P19-1427} {Beyond {BLEU}:training
  neural machine translation with semantic similarity}.
\newblock In \emph{Proceedings of the 57th Annual Meeting of the Association
  for Computational Linguistics}, pages 4344--4355, Florence, Italy.
  Association for Computational Linguistics.

\bibitem[{Wubben et~al.(2012)Wubben, van~den Bosch, and
  Krahmer}]{wubben-etal-2012-sentence}
Sander Wubben, Antal van~den Bosch, and Emiel Krahmer. 2012.
\newblock \href {https://aclanthology.org/P12-1107} {Sentence simplification by
  monolingual machine translation}.
\newblock In \emph{Proceedings of the 50th Annual Meeting of the Association
  for Computational Linguistics (Volume 1: Long Papers)}, pages 1015--1024,
  Jeju Island, Korea. Association for Computational Linguistics.

\bibitem[{Xu et~al.(2015)Xu, Callison-Burch, and Napoles}]{Xu-EtAl:2015:TACL}
Wei Xu, Chris Callison-Burch, and Courtney Napoles. 2015.
\newblock Problems in current text simplification research: New data can help.
\newblock \emph{Transactions of the Association for Computational Linguistics
  (TACL)}.

\bibitem[{Xu et~al.(2016)Xu, Napoles, Pavlick, Chen, and
  Callison-Burch}]{xu-etal-2016-optimizing}
Wei Xu, Courtney Napoles, Ellie Pavlick, Quanze Chen, and Chris Callison-Burch.
  2016.
\newblock \href {https://doi.org/10.1162/tacl_a_00107} {Optimizing statistical
  machine translation for text simplification}.
\newblock \emph{Transactions of the Association for Computational Linguistics},
  4:401--415.

\bibitem[{Xu et~al.(2012)Xu, Ritter, Dolan, Grishman, and
  Cherry}]{xu-etal-2012-paraphrasing}
Wei Xu, Alan Ritter, Bill Dolan, Ralph Grishman, and Colin Cherry. 2012.
\newblock \href {https://aclanthology.org/C12-1177} {Paraphrasing for style}.
\newblock In \emph{Proceedings of {COLING} 2012}, pages 2899--2914, Mumbai,
  India. The COLING 2012 Organizing Committee.

\bibitem[{Zhang et~al.(2020)Zhang, Kishore, Wu, Weinberger, and
  Artzi}]{zhang2020bertscore}
Tianyi Zhang, Varsha Kishore, Felix Wu, Kilian~Q. Weinberger, and Yoav Artzi.
  2020.
\newblock {BERTScore}: Evaluating text generation with {BERT}.
\newblock In \emph{International Conference on Learning Representations}.

\bibitem[{Zhang and Lapata(2017)}]{zhang-lapata-2017-sentence}
Xingxing Zhang and Mirella Lapata. 2017.
\newblock \href {https://doi.org/10.18653/v1/D17-1062} {Sentence simplification
  with deep reinforcement learning}.
\newblock In \emph{Proceedings of the 2017 Conference on Empirical Methods in
  Natural Language Processing}, pages 584--594, Copenhagen, Denmark.
  Association for Computational Linguistics.

\bibitem[{Zhao et~al.(2019)Zhao, Peyrard, Liu, Gao, Meyer, and
  Eger}]{zhao-etal-2019-moverscore}
Wei Zhao, Maxime Peyrard, Fei Liu, Yang Gao, Christian~M. Meyer, and Steffen
  Eger. 2019.
\newblock \href {https://doi.org/10.18653/v1/D19-1053} {{M}over{S}core: Text
  generation evaluating with contextualized embeddings and earth mover
  distance}.
\newblock In \emph{Proceedings of the 2019 Conference on Empirical Methods in
  Natural Language Processing and the 9th International Joint Conference on
  Natural Language Processing (EMNLP-IJCNLP)}, pages 563--578, Hong Kong,
  China. Association for Computational Linguistics.

\bibitem[{Zhu et~al.(2010)Zhu, Bernhard, and
  Gurevych}]{zhu-etal-2010-monolingual}
Zhemin Zhu, Delphine Bernhard, and Iryna Gurevych. 2010.
\newblock \href {https://aclanthology.org/C10-1152} {A monolingual tree-based
  translation model for sentence simplification}.
\newblock In \emph{Proceedings of the 23rd International Conference on
  Computational Linguistics (Coling 2010)}, pages 1353--1361, Beijing, China.
  Coling 2010 Organizing Committee.

\end{thebibliography}
\bibliographystyle{acl_natbib}

\appendix

\clearpage

\section{\metricname~Metric - Implementation Details}
\label{app:metric_impl}

We leverage \textit{Transformers} and \textit{PyTorch Lightning} libraries to implement the metric. We used RoBERTa$_{Large}$ model \cite{Liu2019RoBERTaAR} as the encoder. We fine-tuned the metric with the style-adaptive loss for 20 epochs and selected the checkpoint with the best validation loss. We used the ten human references from ASSET to calculate metric scores while training \metricname~using \datasetnameFirst. We used a batch size of 8, Adam optimizer with a learning rate of 3e-05 for the encoder and 1e-05 for other layers. We use dropout of 0.15 and freeze the encoding layers for 1 epoch. We train the model on one Quadro RTX 6000 GPU with 25GB RAM, which takes around 3 hours.

\section{Incorporating Evaluation Metrics into Training and Inference - Implementation Details}
\label{app:mrt_implementation}

We implement the controllable sentence simplification system proposed by \citet{sheang-saggion-2021-controllable}, a T5-base model. The input sentence is prepended with four control tokens, namely character length ratio, dependency tree depth ratio, character-level Levenshtein similarity, and inverse frequency ratio,  to control various aspects of the generated simplification. During training, each control value is calculated using the corresponding training pair and discretized into bins of width 0.05. During inference, we set the control tokens to the average value of the training set. We used 0.9 for the length ratio and 0.75 for the rest. We also fine-tune the controllable T5-3B and T5-11B versions using the same approach.

We use the Hugging Face Transformers library\footnote{\url{https://huggingface.co/}} for implementing the base model and the MRT framework. We first train the model for 5 epochs using MLE loss and then fine-tune it for 5 epochs using MRT. We use Adam optimizer with a learning rate of 1e-4, linear learning rate warmup of 4k steps, weight decay of 0.1, epsilon of 1e-8, and batch size of 16. During MRT, we use a beam size of 8 to generate candidates. The rest of the parameters are left with default values from the library. During inference, we use a beam size of 10. Our models are trained on 2 A40 GPUs with 45GB RAM for 48 hours.

We used the code released by \citet{fernandes-etal-2022-quality}\footnote{\url{https://github.com/deep-spin/qaware-decode}} for the MBR framework. We generated candidates using beam search of size 100 and selected the top 100 candidates for reranking. We used the above T5 model fine-tuned using MLE to generate the candidates. The rest of the parameters are left with default values from the library.

\section{Systems in \textsc{SimpEval}~Datasets}
\label{app:systems}

\vspace{-.1cm}
\noindent \paragraph{Data-driven Neural Models.} 
We use ten supervised models: (i) three fine-tuned T5 \cite{raffel2020exploring} models of various sizes, namely T5-base, T5-large, and T5-3B, (ii) a controllable T5-base model that prepends tokens to the input to control the lexical and syntactic complexity of the output \cite{sheang-saggion-2021-controllable}, (iii) two BART \cite{lewis-etal-2020-bart} models that also use control tokens where one is trained on Wikipedia \cite{martin-etal-2020-controllable}, and the other is fine-tuned on a combination of Wikipedia and web-mined paraphrases \cite{martin-etal-2022-muss}, (iv) a BERT-initialized Transformer \cite{maddela-etal-2021-controllable}, (v) a BiLSTM edit-based approach that first generates the edit operations, and then the simplification \cite{dong-etal-2019-editnts}, (vi) a BiLSTM that directly generates simplifications using reinforcement learning \cite{zhang-lapata-2017-sentence}, and (vii) a vanilla BiLSTM model. In addition, we also include one unsupervised model and one semi-supervised model by \citet{surya-etal-2019-unsupervised} that uses an auto-encoder with adversarial and denoising losses. 

\vspace{-.1cm}
\noindent \paragraph{Few-shot Methods.} We include simplifications generated by GPT-3.5\footnote{Specifically, we use the \modelfont{text-davinci-003} model which is the most recent variant with 175B parameters.} under zero-shot and 5-shot settings (prompts are provided in Appendix \ref{app:gpt3_setup}).

\vspace{-.1cm}
\noindent \paragraph{Data-driven Statistical Methods.} 
We incorporate two systems that applied statistical machine translation (MT) approaches to text simplification: (i) a phrase-based MT system that reranks the outputs based on their dissimilarity with the input \cite{wubben-etal-2012-sentence}, and (ii) a syntactic MT system that uses paraphrase rules for lexical simplification \cite{xu-etal-2016-optimizing}.

\vspace{-.1cm}
\noindent \paragraph{Rule-based Methods.}  
\citet{dhruv-acl-2020} iteratively generates candidates using rules and ranks them with a linguistic scoring function.

\vspace{-.1cm}
\noindent \paragraph{Hybrid Methods.} 
We utilize three hybrid systems that combine linguistic rules with data-driven methods: (i) \citet{narayan-gardent-2014-hybrid} uses semantic structure to predict sentence splitting and paraphrases with a phrase-based MT system, (ii) \citet{sulem-etal-2018-simple} performs splitting and deletion using linguistic rules and paraphrasing using a BiLSTM, and (iii)  \citet{maddela-etal-2021-controllable} generates candidates with different amounts of splitting and deletion and then paraphrases the best candidate with a BERT-initialized Transformer.

\vspace{-.1cm}
\noindent \paragraph{Naive Baseline Methods.} 
Existing metrics are biased towards conservative systems  because their outputs are generally fluent and exhibit high lexical overlap with the input/reference \cite{pang-gimpel-2019-unsupervised, krishna-etal-2020-reformulating}. We add two conservative systems that are challenging for automatic metrics: (i) a system that always copies the input and (ii) a content-preserving but nonsensical system that scrambles 5\% of the input words. 

\vspace{-.1cm}
\noindent \paragraph{Humans.} We also add human-written simplifications using different instructions from ASSET \cite{fern2020asset} and TurkCorpus \cite{xu-etal-2016-optimizing}, two widely used evaluation benchmarks for sentence simplification, and an auto-aligned one from the SimpleWiki \cite{kauchak-2013-improving}. 

\section{Implementation Details for Simplification Systems}
\label{app:implementation_detail}

\subsection{T5 Setup}
\label{app:t5_setup}
We use the Hugging Face Transformers library\footnote{\url{https://huggingface.co/}}. We fine-tune T5-base, T5-large, and T5-3B on 4 A40 GPUs of a total batch size of 64 for 20 epochs (10.4K steps), 16 epochs, and 8 epochs, respectively. We use a learning rate of 3e-4. We save checkpoints every 5K steps and select the best one by performing a manual inspection on a set of 60 simplifications from the development set, resulting in 80K, 60K, and 25K steps for T5-base, T5-large, and T5-3B respectively, which are in a similar range to FLAN-T5 \cite{chung2022scaling}. We use AdamW \cite{loshchilov2017decoupled} as the optimizer.

\subsection{GPT-3.5 Setup}
\label{app:gpt3_setup}
\paragraph{Hyperparameters.}
We use the text-davinci-003 GPT-3.5 model from OpenAI API\footnote{\url{https://beta.openai.com/}}. To generate simplification, we use the following hyperparameters: temperature=1 and top-p=0.95.

\paragraph{Prompts.}
We use the instruction from ASSET \cite{alva-manchego-etal-2021-un} to prompt GPT-3.5.

\paragraph{Zero-shot setting:}
\textit{Please rewrite the following complex sentence in order to make it easier to understand by non-native speakers of English. You can do so by replacing complex words with simpler synonyms (i.e. paraphrasing), deleting unimportant information (i.e. compression), and/or splitting a long complex sentence into several simpler ones. The final simplified sentence needs to be grammatical, fluent, and retain the main ideas of its original counterpart without altering its meaning. \\
\\
Input: \{input\} \\
Output:}

\paragraph{5-shot setting:}
\textit{
Please rewrite the following complex sentence in order to make it easier to understand by non-native speakers of English. You can do so by replacing complex words with simpler synonyms (i.e. paraphrasing), deleting unimportant information (i.e. compression), and/or splitting a long complex sentence into several simpler ones. The final simplified sentence needs to be grammatical, fluent, and retain the main ideas of its original counterpart without altering its meaning.\\
\\
Examples:\\
Input: \{random sampled from ASSET\}\\
Output: \{random sampled human reference of the Input from ASSET\}\\
\\
Input: \{random sampled from ASSET\}\\
Output: \{random sampled human reference of the Input from ASSET\}\\
\\
Input: \{random sampled from ASSET\}\\
Output: \{random sampled human reference of the Input from ASSET\}\\
\\
Input: \{random sampled from ASSET\}\\
Output: \{random sampled human reference of the Input from ASSET\}\\
\\
Input: \{random sampled from ASSET\}\\
Output: \{random sampled human reference of the Input from ASSET\}\\
\\
\\
Please rewrite the following complex sentence in order to make it easier to understand by non-native speakers of English. You can do so by replacing complex words with simpler synonyms (i.e. paraphrasing), deleting unimportant information (i.e. compression), and/or splitting a long complex sentence into several simpler ones. The final simplified sentence needs to be grammatical, fluent, and retain the main ideas of its original counterpart without altering its meaning.\\
\\
Input: \{input\} \\
Output:}

\clearpage 

\onecolumn

\section{\metricname~metric - Additional Results}
\label{app:additional_results}

\setlength{\tabcolsep}{5pt}
\begin{table*}[ht!]
\small
\centering
\begin{tabular}{lccccccccc}
\toprule 
& \multicolumn{3}{c}{\textbf{\datasetname $_{2022}$}} 
& \multicolumn{3}{c}{\textbf{\textsc{WIKI-DA}}} 
& \multicolumn{3}{c}{\textbf{\textsc{Newsela-Likert}}} \\
\cmidrule(lr){2-4} \cmidrule(lr){5-7} \cmidrule(lr){8-10}
& $\mathbf{\mathlarge{\tau}_{para}}$ & 
$\mathbf{\mathlarge{\tau}_{spl}}$ & $\mathbf{\mathlarge{\tau}_{all}}$ &
\textbf{Fluency} & \textbf{Meaning} & \textbf{Simplicity} & 
\textbf{Fluency} & \textbf{Meaning} & \textbf{Simplicity }\\ 
\midrule
FKGL & -0.556 & -0.31 & -0.356  & 0.054 & 0.145 & 0.001 
 & 0.193 & 0.306 & -0.051 \\
BLEU & 0.048 & -0.054 & -0.033 & 0.460 & 0.622 & 0.438
& 0.332 & 0.261 & 0.118 \\
SARI &  0.206 & 0.140 & 0.149 & 0.335 & 0.534 & 0.366
& 0.234 & 0.124 & 0.094 \\
BERTScore & 0.238 & 0.093 & 0.112 & 0.636 & \underline{0.682} & 0.614  & 0.384 & 0.274 & 0.215 \\
\midrule
\multicolumn{10}{l}{\textbf{\textit{Original \metricname}}} \\
\metricname$_{all}$ & 0.333 & 0.233 & 0.241 & \underline{0.823} & 0.665 & 0.715 & \underline{0.654} & \underline{0.471} & 
{0.336} \\
\metricname$_{k=1}$ & \textbf{0.460} & \underline{0.295} & \underline{0.307}  & 0.813 & 0.658 & 0.692  & 0.649 & 0.449 & 0.321 \\
\metricname$_{k=3}$ & \underline{ 0.429} & \textbf{0.333} & \textbf{0.331}  & \textbf{0.843} & \textbf{0.684} & \underline{0.735} & 0.646 & 0.437 & \underline{0.353} \\
\midrule
\multicolumn{10}{l}{\textbf{\textit{Rescaled \metricname}}} \\
\metricname$_{all}$ & 0.333 & 0.233 & 0.241 & 0.816 & 0.662 & 0.733 & \textbf{0.655} & \textbf{0.477} & {0.343}\\
\metricname$_{k=1}$ & \textbf{0.460} & \underline{0.295} & \underline{0.307}  & 0.796 & 0.647 & 0.721  & 0.633 & 0.444 & 0.328 \\
\metricname$_{k=3}$ & \underline{ 0.429} & \textbf{0.333} & \textbf{0.331}  & 0.807 & 0.668 & \textbf{0.749} & 0.624 & 0.428 & \textbf{0.359} \\
\bottomrule
\end{tabular}
\caption{Metric evaluation results on {\sc SimpEval}$_{2022}$, \textsc{WIKI-DA} \cite{alva-manchego-etal-2021-un}, and \textsc{Newsela-Likert} \cite{maddela-etal-2021-controllable} human ratings datasets. We include the results for both original and rescaled versions of \metricname. $\mathbf{\mathlarge{\tau}_{para}}$, $\mathbf{\mathlarge{\tau}_{spl}}$, and $\mathbf{\mathlarge{\tau}_{all}}$ represent  the Kendall Tau-like correlation for paraphrase-focused, split-focused, and all simplifications respectively. We report the Pearson correlation coefficients along three dimensions for \textsc{WIKI-DA} and \textsc{Newsela-Likert}. The best values are marked in \textbf{bold} and the second best values are \underline{underlined}. As Kendall Tau measures pairwise rankings, we see the same results for the original and rescaled versions of \metricname.}
\label{table:metric_eval_results_simpeval_appendix}
\end{table*}

\clearpage

\section{Annotation Interface}
\label{sec:interface}
\subsection{Step - 1: System Output Categorization.}

\begin{minipage}{\textwidth}
    \centering
    \includegraphics[width=0.75\textwidth]{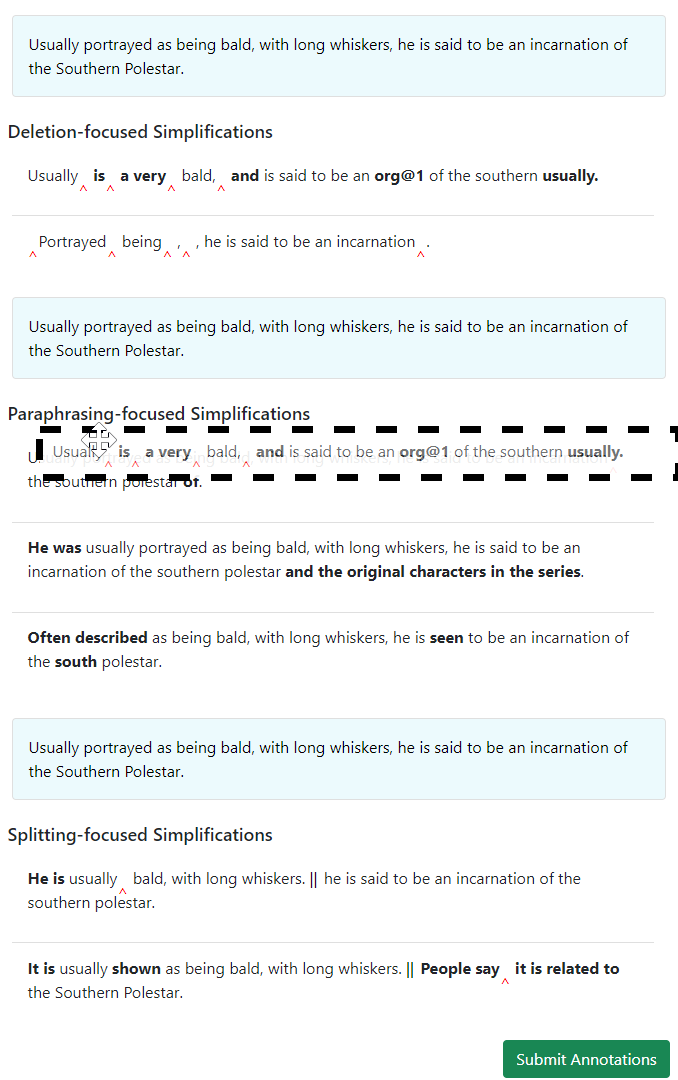}
    \captionof{figure}{Annotation interface for categorizing system outputs. The outputs can be moved up and down or to other categories.}
    \label{fig:categorization}
\end{minipage}

\clearpage

\subsection{Step - 2: Highlighting System Edits.}

\noindent\begin{minipage}{\textwidth}
    \centering
    \includegraphics[width=0.45\textwidth]{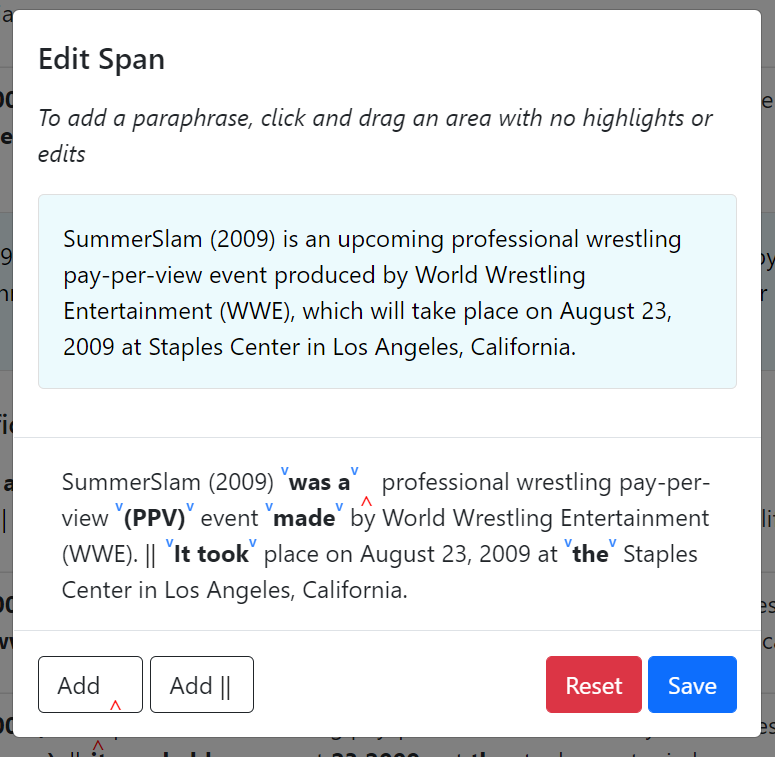}
    \captionof{figure}{Span-fixing interface provided by annotators to fix the highlighted changes between the original and simplified sentences. Annotators could click and drag to modify the bounds of a paraphrase label or highlight text to add a paraphrase level. The deletion (``$\wedge$'') and split (`` \textbf{||} '') can also be dragged and can be added using buttons on the bottom left of the modal.}
    \label{fig:span_fixing}
    
    \vspace{15pt}
    
    \centering
    \includegraphics[width=0.6\textwidth]{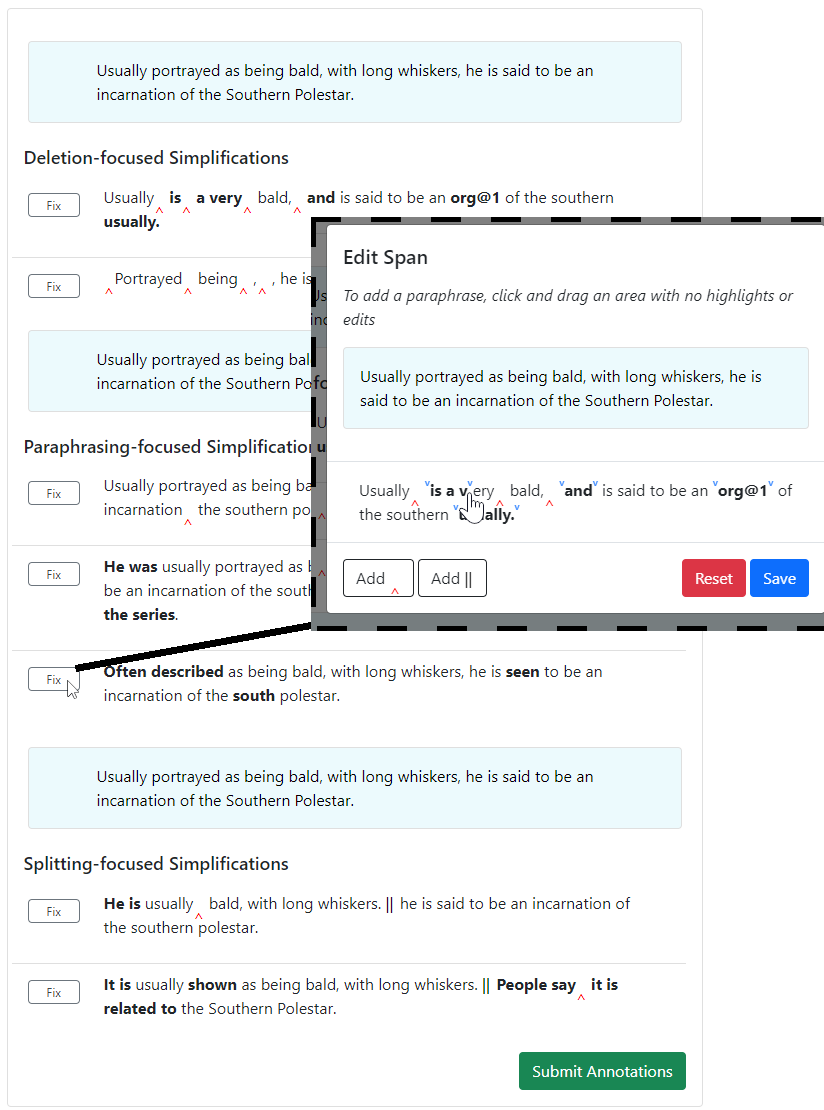}
    \captionof{figure}{Annotation interface for span fixing. Clicking on ``Fix'' button opens up a pop up window.}
    \label{fig:span_fixing_overall}
\end{minipage}

\clearpage

\subsection{Step - 3: Rating and Ranking System Outputs.}
\begin{minipage}{\textwidth}
    \small
    \centering
    \includegraphics[width=\textwidth]{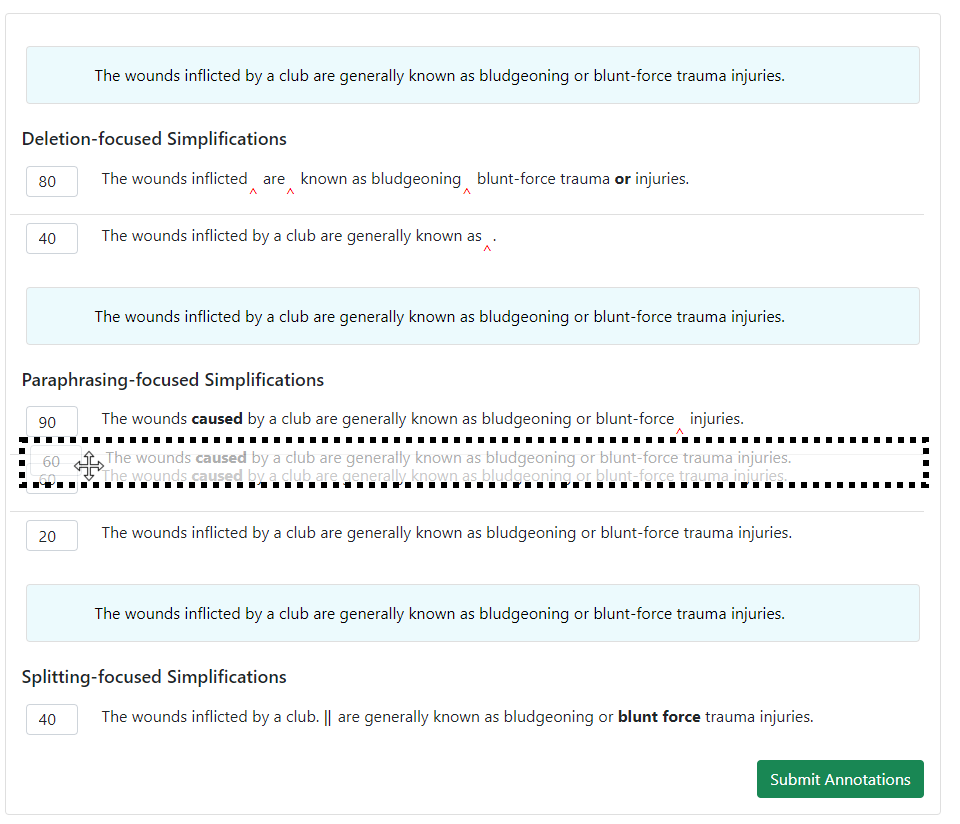}
    \captionof{figure}{Sentence ranking and rating interface provided to annotators. The annotator can enter the ratings for each sentence and is able to re-order sentences by clicking and dragging their mouse.}
    \label{fig:dragging}
\end{minipage}

\clearpage

\subsection{Annotation Instructions}
\label{sec:instructions}

\begin{minipage}{\textwidth}
    \centering
    \includegraphics[width=0.8\textwidth]{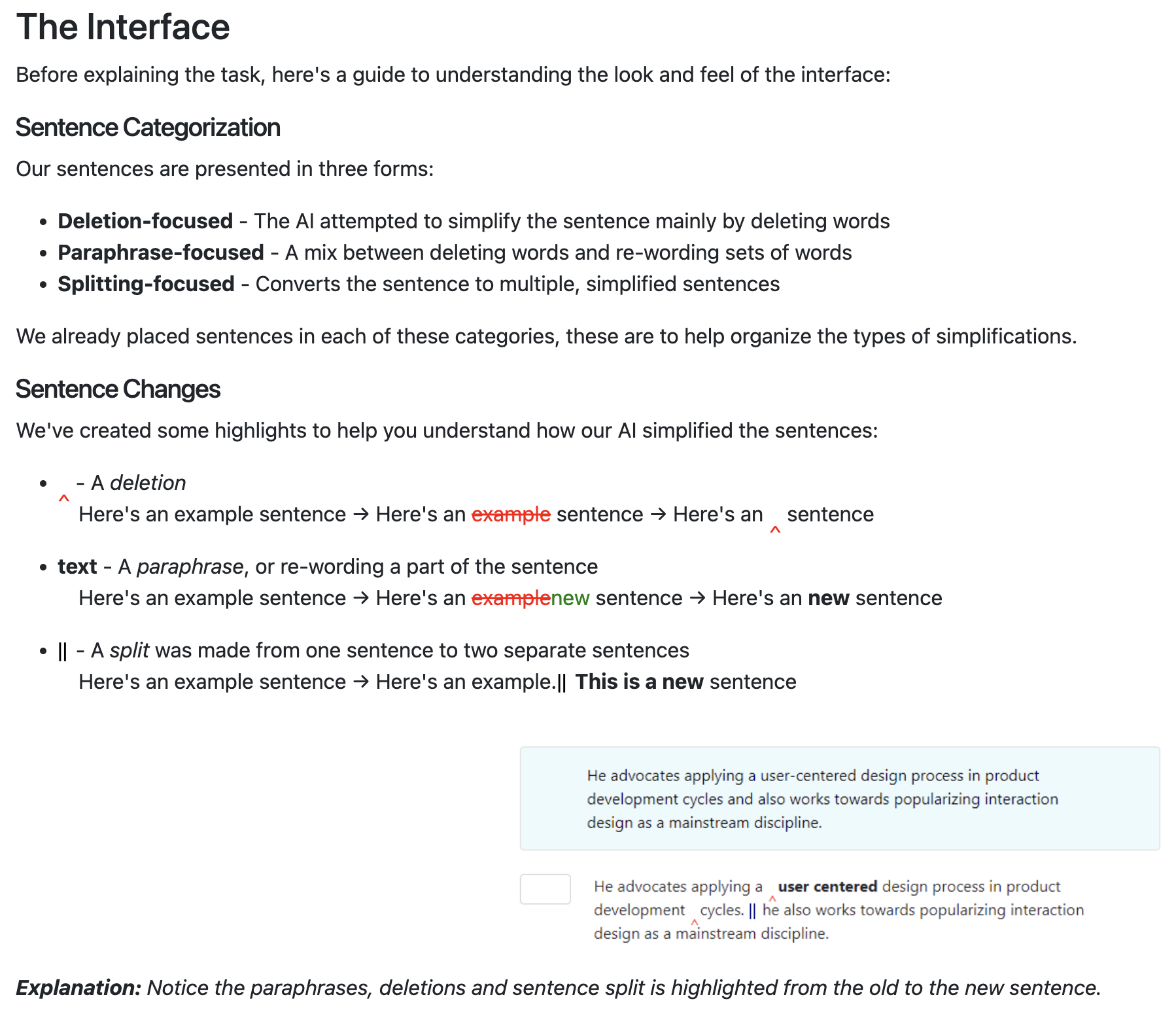}
    \captionof{figure}{Instructions for the overall task.}
    \label{fig:task_instruction}
\end{minipage}

\clearpage

\begin{minipage}{\textwidth}
    \centering
    \includegraphics[width=0.8\textwidth]{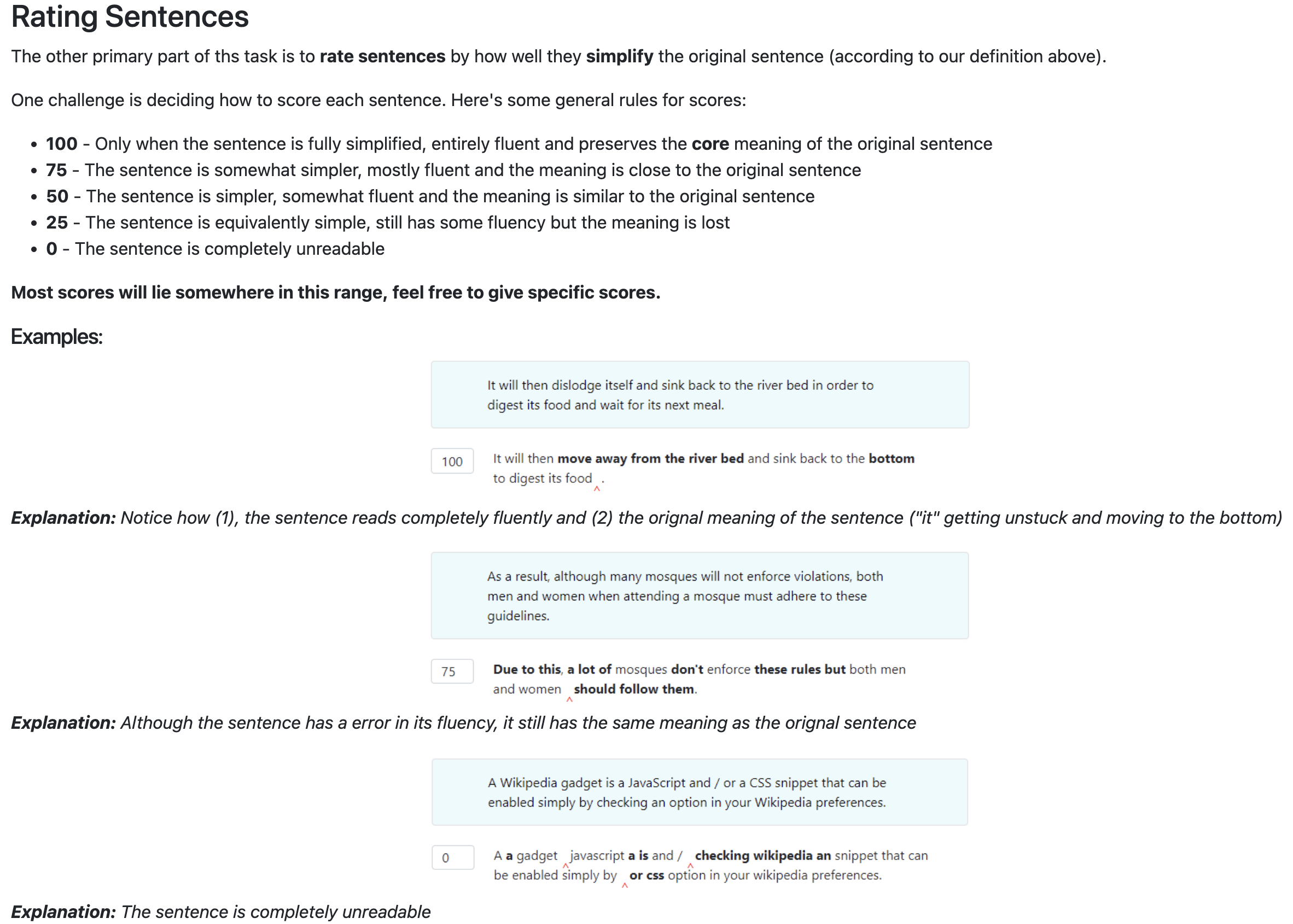}
    \captionof{figure}{Instructions for ranking and rating.}
    \label{fig:rating}
\end{minipage}

\clearpage

\end{document}